\documentclass{article}
\usepackage[utf8]{inputenc}

\usepackage[english]{babel}

\usepackage[letterpaper,top=2cm,bottom=2cm,left=3cm,right=3cm,marginparwidth=1.75cm]{geometry}
\usepackage[ruled,vlined]{algorithm2e}
\usepackage{amsmath, amsfonts,amssymb}
\DeclareMathOperator*{\argmax}{arg\,max}
\usepackage{multirow, multicol, makecell, booktabs, graphicx}
\usepackage{caption}
\usepackage{subcaption}
\usepackage{lscape}
\usepackage{float}
\usepackage{subcaption, natbib, authblk, footnote, url}
\usepackage[acronym,nogroupskip,nopostdot]{glossaries}
\usepackage{xltabular}
\usepackage[colorlinks=true, allcolors=blue]{hyperref}
\usepackage[short]{optidef}
\usepackage{subfiles, nameref, hyperref}
\hypersetup{
     colorlinks = true,
     citecolor  = blue,
     linkcolor = blue,
     urlcolor = blue
}
\usepackage{cleveref}
\usepackage{geometry, algorithm2e}
\RestyleAlgo{ruled}
\usepackage{setspace}
\usepackage[title]{appendix}
\usepackage[colorinlistoftodos]{todonotes}

\title{All You Need is an Improving Column: Enhancing Column Generation for Parallel
Machine Scheduling via Transformers}

\author{Amira Hijazi}
\affil{NSF AI Institute for Advances in Optimization (AI4OPT), \protect \\ Georgia Institute of Technology, Atlanta, GA}

\author{Osman Ozaltin, Reha Uzsoy}
\affil{Edward P. Fitts Department of Industrial and Systems Engineering, \protect \\ North Carolina State University}
\date{}

\begin{document}
\maketitle

\begin{abstract}
We present a neural network-enhanced column generation (CG) approach for a parallel machine scheduling problem. The proposed approach utilizes an encoder-decoder attention model, namely the transformer and pointer architectures, to develop job sequences with negative reduced cost and thus generate columns to add to the master problem. By training the neural network offline and using it in inference mode to predict negative reduced costs columns, we achieve significant computational time savings compared to dynamic programming (DP). Since the exact DP procedure is used to verify that no further columns with negative reduced cost can be identified at termination, the optimality guarantee of the original CG procedure is preserved. For small to medium-sized instances, our approach achieves an average 45\% reduction in computation time compared to solving the subproblems with DP. Furthermore, the model generalizes not only to unseen, larger problem instances from the same probability distribution but also to instances from different probability distributions than those presented at training time. For large-sized instances, the proposed approach achieves an 80\% improvement in the objective value in under 500 seconds, demonstrating both its scalability and efficiency.\\
\noindent\emph{\textbf{Keywords}:
Column Generation, Machine Learning, Transformers, Parallel Machine Scheduling
}\\
\end{abstract}

\clearpage
\section{Introduction}

\noindent Parallel machine scheduling has applications in many manufacturing and service systems~\citep{weng2001unrelated, hall2000parallel, baesler2014multiobjective}. The problem involves allocating and sequencing a set of jobs $N = \{1, 2, ..., n\}$ on a set of parallel machines $M = \{1,2,...,m\}$ where each job is assigned to only one machine and each machine can process only one job at a time. An instance of the parallel machine scheduling problem is defined by the job attributes (processing time, weight), machine environment (identical, proportional, or unrelated), and a performance measure such as makespan~\citep{mokotoff2001parallel}, total weighted completion time~\citep{chen1999solving}, or total tardiness~\citep{yalaoui2002parallel}. We consider the problem of minimizing the total weighted completion time on unrelated parallel machines~\citep{lawler1993sequencing}. In this problem, each job $j \in N$ has a weight $w_j$ and a processing time $p_{jk}$ on machine $k \in M$. The job processing times on different machines are independent of each other, hence \textit{unrelated} machines. 

A common solution approach for this problem is column generation~\citep{wilhelm2001technical, lubbecke2005selected}. Each iteration of column generation involves solving a linear program, called the restricted master problem (RMP), restricted to a subset of the variables. Additional columns that will improve the objective function value of the RMP are generated as needed by solving an application-specific pricing subproblem. If such columns are found, they are added to the restricted LP, which is re-optimized. Iterations continue until no further improving columns can be found. The optimal solution of the restricted master problem must then be processed in some way to obtain a feasible integer solution. The pricing subproblem for the scheduling problem we consider is a single-machine scheduling problem that can be solved using a pseudo-polynomial time dynamic programming (DP) algorithm, which becomes computationally expensive as the instance size increases. 

\begin{figure}
   \caption{\texttt{CG-DP} vs \texttt{CG-NN-DP} }

    \includegraphics[width=0.4\linewidth]{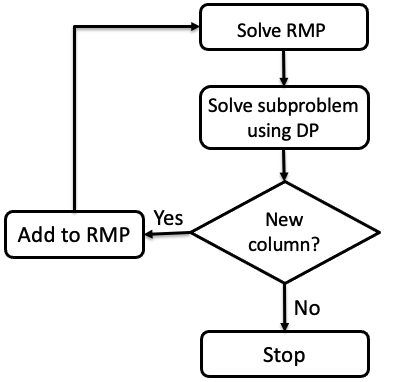}\hfill
    \includegraphics[width=0.6\linewidth]{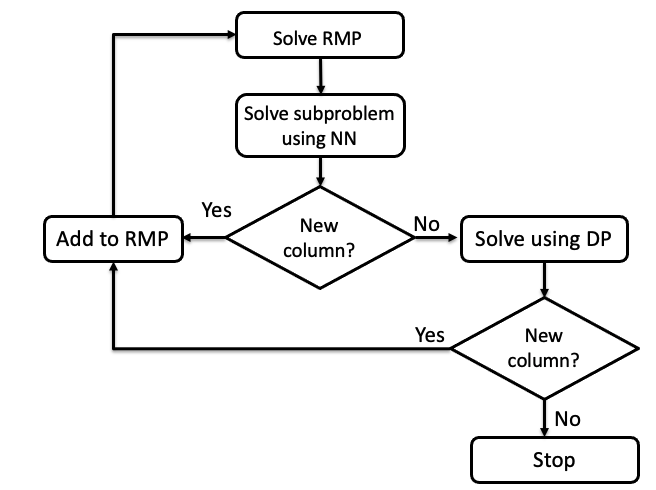}
    \label{fig:CG}
\end{figure}

In this paper, we leverage the capabilities of machine learning (ML), specifically neural networks, to rapidly approximate the solution of the pricing subproblem to obtain improving columns with negative reduced cost. Flowcharts of the traditional CG approach with DP (\texttt{CG-DP}) and our proposed CG approach (\texttt{CG-NN-DP}) are illustrated in Figure ~\ref{fig:CG}. In our proposed approach, if the column generated by the NN has a negative reduced cost, we add it to the RMP and reoptimize. If no column with a negative reduced cost is predicted by the NN, we solve the subproblem using DP to confirm that no further improving columns exist. Since the neural network is trained offline and used in inference mode to predict a job sequence with a negative reduced cost, the savings in computation time over the DP are significant. Since the exact DP procedure is used to verify that no further columns with negative reduced cost can be identified at termination, the optimality guarantee of the original CG procedure is preserved. 
The proposed neural network uses the transformer architecture with a pointer layer. Transformers are encoder-decoder attention-based neural networks designed for sequential data tasks that have been successfully used in language translation and generating graph structures~\citep{yun2020graph}. In the combinatorial optimization domain, transformers have been used for solving the Traveling Salesman Problem (TSP)~\citep{kool2018attention}, Vehicle Routing Problem~\citep{wu2024neural}, Knapsack Problem~\citep{lee2024attention}, and Job Shop Scheduling~\citep{ yildiz2022reinforcement}. The main advantages of transformers over Recurrent Neural Networks (RNN) and Long Short-Term Memory Networks (LSTM) networks arise from enhanced modeling of the interactions between the different input vectors representing job and machine parameters and improved training efficiency due to the attention mechanism. 

Our initial experiments using a transformer architecture alone were only able to achieve 45\% validation accuracy and failed to generate negative reduced cost columns when used within the CG in the testing phase, so we added a pointer layer to the transformer to improve the learning. Pointer networks were introduced by~\citet{vinyals2015pointer} based on sequence-to-sequence networks~\citep{sutskever2014sequence} to learn the conditional probability $p(C^P|P)$ of an output sequence $C^P$ with elements corrresponding to positions in an input set $P$. They have been used successfully in approximating solutions to several combinatorial optimization problems, such as finding the convex hull and Delaunay triangulation~\citep{vinyals2015pointer}, the Traveling Salesman Problem~\citep{stohy2021hybrid, ma2019combinatorial}, and the selection of branching variables in branch and bound algorithms ~\citep{wang2024learning}. Pointer networks have demonstrated their ability to generalize to instances larger than those in the training sets, making them well-suited for combinatorial optimization problems like parallel machine scheduling.

\subsection{Literature Review}

\textbf{Parallel Machine Scheduling.}
Most variants of the parallel machine scheduling problem, including that addressed in this paper, are NP-hard in the strong sense, rendering them computationally challenging for commercial solvers~\citep{pfund2004survey, li2009non}. A variety of
exact and approximation solution methods have been developed as a result. \citet{phillips1994task} proposed the first polynomial-time approximation algorithm for minimizing the total weighted completion time on unrelated parallel machines. \citet{azizoglu1999scheduling} developed an exact branch and bound algorithm capable of solving small instances (up to three machines and 20 jobs). \citet{chen1999solving} developed a branch and price algorithm, which forms the basis of the approach in this paper. The extensive literature on parallel machine scheduling problems is reviewed by~\cite{ mokotoff2001parallel, li2009non} and ~\cite{durasevic2023heuristic}.\\
\textbf{ML for Combinatorial Optimization.}
Machine learning techniques have been applied to combinatorial optimization problems through two distinct strategies. The first involves using ML tools to enhance decision-making within optimization algorithms themselves, either through supervised learning or reinforcement learning. \citet{khalil2016learning} proposed a supervised learning framework for variable branching in mixed integer programming imitating the strong branching rule.  \cite{tang2020reinforcement} utilized a reinforcement learning framework to learn a cut selection policy for the cutting plane algorithm, which is used in branch-and-cut solvers. \citet{rajabalizadeh2024solving} used classification techniques to approximate the optimal value of cut-generating linear programs.
The second strategy involves using ML tools to approximate solutions to difficult optimization problems, often referred to as "optimization proxies." \citet{bresson2021transformer} trained a reinforcement learning agent to solve the TSP problem for instances as large as 100 nodes while~\citet{hertrich2023provably} developed an RNN to predict solutions to the knapsack problem.~\cite{ojha2023optimization} developed an optimization proxy for outbound load planning in the parcel service industry using supervised learning. We refer the readers to survey papers by~\cite{bengio2021machine, mazyavkina2021reinforcement, kotary2021end, zhang2023survey, cappart2023combinatorial}, and ~\cite{fan2024artificial}. 

\noindent \textbf{ML for Column Generation.} Several recent studies have applied ML methods to Column Generation.~\citet{sugishita2024use}  utilized ML to generate initial dual values to warm-start the column generation procedure for unit commitment problem.~\citet{morabit2021machine} proposed a supervised Graph Neural Network (GNN) approach for selecting the most promising column from a set of available columns obtained by solving the pricing problem. They demonstrate the effectiveness of their approach on the Vehicle and Crew Scheduling Problem and the Vehicle Routing Problem with Time Windows, obtaining reductions in computing time of up to $30\%$.

Instead of relying on a time-consuming exact method, \citet{chi2022deep} adopt deep reinforcement learning with GNN to optimize the column selection based on the maximum expected future reward.  Their experimental results demonstrate significant reductions in the number of column generation (CG) iterations for the Vehicle Routing Problem with Time Windows compared to the traditional greedy column selection policy.
In contrast, \citet{shen2022enhancing} used supervised machine learning techniques to construct a pricing heuristic for the graph coloring problem. A linear Support Vector Machine is used to predict the optimal solution for the pricing problem, which is then used  to guide an efficient sampling technique that generates many high-quality columns.  The findings indicate a significant decrease in the number of pricing iterations needed, resulting in faster convergence and enhanced computational efficiency.

\subsection{Contributions and Organization}

In this paper, we train a neural network (NN) to speed up the CG solution time by predicting the optimal solution to the pricing subproblem instead of relying solely on  dynamic programming. The contributions of this paper are as follows: 
\begin{enumerate}
    \item We propose a ML approach to predict the optimal solution of a difficult combinatorial scheduling problem, namely the single machine scheduling subproblem. 
    \item The proposed approach is integrated with the CG approach to obtain a new column for each pricing subproblem by predicting a candidate sequence of jobs for the associated machine instead of using DP.  If no column with a negative reduced cost is predicted by the neural network, we solve the subproblem using dynamic programming to confirm that no further improving columns exist. This approach achieves significant time reduction and improved convergence compared to the greedy CG with DP, while preserving the optimality guarantee of the original CG procedure. 
    \item We train a single model that considers different instance sizes, including different numbers of machines and jobs, and demonstrate its ability to generalize performance beyond the instance sizes and parameter distributions in the training set.
\end{enumerate}

Section~\ref{Problem Formulation}  presents the parallel machine scheduling problem formulation, and  Section~\ref{framework section} introduces the different solution methods.  Section~\ref{Data Generation} includes details about the data generation pipeline and Section~\ref{experiments} describes the experiment settings. Section~\ref{Results} reports the computational results, and Section~\ref{Conclusion} concludes the paper. 

\section{Problem Formulation} 
\label{Problem Formulation}
We consider the problem of minimizing the total weighted completion time on unrelated parallel machines, denoted by  $R\|\sum w_jC_j$ in the standard three-field notation~\citep{pinedo2012scheduling}, where $C_j$ denotes the completion time of job $j$ in a schedule and $w_j$ the weight of job $j$. The objective is to assign jobs to machines and sequence the assigned jobs on each machine to minimize the total weighted completion time. All jobs are simultaneously available with no precedence constraints or setup times. Our point of departure is the branch and price approach of~\citet{chen1999solving}, who use Dantzig-Wolfe decomposition~\citep{dantzig1960decomposition} to decompose their original integer programming formulation into a set partitioning master problem and pricing subproblems associated with each of the machines that are solved using dynamic programming.

Let $M$ denote the set of machines and $\Omega_k$ the set of all possible schedules on machine $k$ in which a subset of jobs is processed on that machine. For a schedule $s_k$ containing $l$ jobs, the job sequence must follow the shortest weighted processing time (SWPT) order~\citep{smith1956various}; $\frac{p_{[1]}}{w_{[1]}} \leq \frac{p_{[2]}}{w_{[2]}} \leq ... \leq \frac{p_{[l]}}{w_{[l]}}$ where $[i]$ denotes the job $j$ with the $i$th smallest value of $\frac{p_{j}}{w_{j}}$. The total weighted completion time of schedule $s_k$ on machine $k$ is given by $f_s^k   = \sum^l_{i=1}w_{[i]}C_{[i]}$.

For each job $j \in N$, let $a_{js}^k = 1$ if job $j$ is processed in schedule $s_k \in \Omega_k$ on machine $k$, and 0 otherwise.
Defining the 0-1 decision variables $y_s^k$ for $k \in M, s \in \Omega^k$ as:
\begin{equation*}
     y_s^k = \begin{cases}
       1, & \text{if schedule }  s \in \Omega^k  \text { is used}\\
        0, & \text{otherwise}
        \end{cases}
\end{equation*}
the set partitioning master problem proposed in~\citet{chen1999solving} is written as\label{SetPartitioning} : 
\begin{align} 
(MP) \quad \min \ &\sum_{k \in M}\sum_{s \in \Omega^k} f_s^k y_s^k \label{obj1} \\
& \sum_{k \in M} \sum_{s \in \Omega^k} a_{js}^k y_s^k = 1   & \forall j \in N \qquad (\pi_j)\label{JobConstraint}\\
&\sum_{s \in \Omega^k} y_s^k \leq 1  & \forall k \in M \qquad (\sigma_k)\label{MachineConstraint} \\
&y_s^k \in \{0, 1\} & \forall s \in \Omega^k, k \in M  \label{integ}
\end{align}
The objective function~\eqref{obj1} minimizes $\sum w_jC_j$ over all machines. Constraints ~\eqref{JobConstraint} ensure that each job is processed in exactly one partial schedule, and~\eqref{MachineConstraint} that at most one schedule is selected for each machine. The LP relaxation of the master problem is obtained by relaxing constraints ~\eqref{integ} such that $0 \leq y_s^k \leq 1$. Constraints~\eqref{MachineConstraint} satisfy the upper bound constraints $y_s^k \leq 1$, making explicit upper bound constraints redundant. 

The \textbf{column generation} approach starts by solving an LP relaxation of the MP with a restricted number of columns, the Restricted Master Problem (RMP), where a column represents a feasible schedule for a given machine. 
A single machine subproblem for each machine $k\in M$ is solved using dynamic programming to find a feasible schedule $s \in \Omega^k$ with negative reduced cost. The reduced cost of a feasible schedule $s \in \Omega^k$ on machine $k \in M$ is given by:
\begin{equation}
     r_s^k = f_s^k - \sum_{ j \in N}a_{js}^k\pi_j - \sigma_k, \label{reducedcostDP}
\end{equation}
where $\pi_j$, $j \in N$ denote the dual variables associated with constraints~\eqref{JobConstraint} and $\sigma_k$, $k \in K$  those associated with constraints ~\eqref{MachineConstraint}.

 If such columns exist, they enter the basis, and the LP relaxation of the RMP is re-optimized until no such columns can be found. If the optimal solution at this point satisfies the integrality property, then the solution is optimal for the original IP problem. However, this is generally not the case. Therefore, column generation is combined with branch and bound to find integer solutions, and this method is called branch and price. In this work, we restrict our focus on column generation.  For small to medium-sized instances, we run the CG until no further columns with a negative reduced cost can be found and then solve the restricted master problem containing all columns present in the final solution to its LP relaxation with integer $y^k_s$, to ensure we have an integer solution, thus obtaining an optimality certificate of the original integer program. For large instances, we run the CG with a time limit so  optimality is not guaranteed. 

\noindent{\textbf{Initial Solution.}} A feasible solution to the RMP is required to initiate the column generation procedure. For this purpose, we explore two different list scheduling heuristics. 
The first randomly selects a subset of jobs, with each job having an equal likelihood of being chosen, and sorts them based on the SWPT rule. The number of times the heuristic is executed per machine ranges from 5 to 100, depending on the instance size.  The second heuristic generates a random permutation of the jobs and sequentially assigns each job to the machine with the minimum total weighted completion time. The heuristic is executed 2,000 times to generate a diverse set of initial solutions. The 20 solutions with the lowest total weighted completion times are selected to initiate the column generation procedure. Despite its potential advantages, we did not observe any significant time improvement overall when using the second heuristic. Therefore, we employ the computationally more efficient first heuristic to initialize the column generation (CG) instances used for training and testing the neural network.

\section{Pricing Subproblem Solution Methods}
\label{framework section}
In this section, we present the dynamic programming approach for the subproblems and the neural network approaches: a transformer neural network and a transformer pointer neural network. 
\subsection{Baseline Solution Method: Dynamic Programming}
The pseudo-polynomial time dynamic programming algorithm in~\citet{chen1999solving} derives from the procedure by~\citet{rothkopf1966scheduling}, which exploits the fact that in any optimal schedule for the $R\|\sum w_jC_j$ problem, jobs on each machine must follow the SWPT sequence. Their algorithm for the pricing subproblem, stated in Algorithm~\ref{Chen et al. Dynamic Programming Algorithm}, defines the recursion function $F(j, t)$ as the minimum objective function value of a schedule containing a subset of jobs $\{1, 2, ..., j\}$ sequenced by the SWPT rule and completed at time $t$. The pseudo-polynomial time complexity of this DP is due to its selecting a subset of jobs for the partial schedule, rendering it NP-hard in the ordinary sense~\citep{garey1979computers}.
\begin{algorithm}
  \caption{Chen et al. Dynamic Programming Algorithm}
  \label{Chen et al. Dynamic Programming Algorithm}
  \SetAlgoLined
  \SetKwInOut{Initialize}{Initialize}
  \SetKwInOut{KwReturn}{Return}
  \Initialize{$F(j,t) = \infty, \text{  } t < 0, j = 0, ..., n$ \\$F(0,t) = 0,  \text{  } t \geq 0 $}
  \For{$j  \in (1,...,n)$}{ 
             \For {$t \in (0,...,P)$}{
                    $F(j,t) = \min \{F(j-1, t- p_j) + tw_j - \pi_j, F(j-1, t) \}$
             }
        }
  \KwReturn{$\min_{j \in N, 0 \leq t \leq P} \{F(j,t)\}$}
\end{algorithm}
The worst-case time complexity of this algorithm is $O(n^2P)$, where $P$ denotes the total processing time of all jobs and $n$ the number of jobs, which can be computationally expensive for large instances with long processing times, motivates our proposal to replace the DP with a machine learning model.

\subsection{Neural Network Approaches}
In the NN framework, an instance  $(X,s)$ of the pricing subproblem corresponds to a single machine scheduling subproblem encountered in one CG iteration on a specific machine $m$, where $s$ is the target schedule for input feature matrix $X = [x_0,x_1,\ldots,x_n]$. 
The first row of the input matrix $X$ is the machine vector: $x_0 = [0, 0, \sigma_m]$, where $\sigma_m$ is the dual variable associated with machine $m$ via constraints~\eqref{MachineConstraint}. Each job is represented by a row $x_j = [p_{jm}, w_j, \pi_j]$, where the job features $p_{jm}, w_j, \pi_j$ denote its processing time on machine $m$, weight, and the value of its associated dual variable via constraints ~\eqref{JobConstraint} in the current solution to the restricted master problem, respectively. The target schedule $s$ consists of a subset of jobs on machine $m$ in SWPT sequence yielding the most negative reduced cost computed by the dynamic program. 

To predict $s$ given input $X$, we utilize a NN model with learnable weights $\theta$. The NN model estimates the probability $p_m(s|X; \theta)$ based on the chain rule:

\begin{center}
      $ p_m(s|X; \theta) = \Pi_{i=1}^n p_m(s_i|s_0, ..., s_{i-1}, X;\theta)$ 
\end{center}
 
The conditional probability $p_m(s_i|s_0,..., s_{i-1}, X; \theta)$ indicates the likelihood of selecting $s_i$ to be in the $i$th schedule position, given the input $X$ and previously selected elements $s_0,..., s_{i-1}$ in the schedule. The predicted schedule $\hat{s}$ is chosen to be the one with the highest probability, i.e., $\hat{s} \in \argmax p(s|X;\theta)$.

\subsubsection{Input Preprocessing}
To prepare the input $X$ for the transformer architecture, we append two special vectors to it indicating the start, denoted by ($\Rightarrow$), and end of the schedule, denoted by ($\Leftarrow$). $\Rightarrow$ is represented by the $1 \times 5$  vector $[0,0,0,1,0]$ and $\Leftarrow$ by $[0,0,0,0,1]$. The job input vector $x_j$ and machine input vector $x_0$ are then modified as: $x_j = [p_{jm}, w_j, \pi_j, 0, 0]$ and $x_0 = [0,0,\sigma_m, 0, 0]$ to ensure consistent dimensions for matrix operations, padding and masking, batch processing, and parallel computation.

The target schedule is also preprocessed to include the machine, $\Rightarrow$, and $\Leftarrow$ elements. 
For example, target schedule $[3, 1, 2]$ is input to the network as  $s = [\Rightarrow, 3, 1, 2, 0, \Leftarrow]$ where $\Rightarrow (s_0)$ indicates the beginning of the schedule, $s_1=3, s_2=1, s_3=2$ the indices of the job vectors, and $0$ the index of the machine vector in the input sequence.  Although we use the machine vector index $0$ after the last job index in the schedule, we still need $\Leftarrow$ to indicate the end of schedule because the machine vector differs between different input instances while $\Leftarrow$ remains constant. Henceforth we refer to $s$ as a schedule and NN output interchangeably. 

\begin{figure}
    \caption{Transformer-Pointer Network with input $X = \{x_0, x_1, x_2, x_3, x_4\}$, and output $\{\Rightarrow, 3, 1, 2, 0, \Leftarrow\}$ from which we have the output schedule $[3,1,2]$. The elements $\Rightarrow$ and $\Leftarrow$ represent the beginning and end of the schedule, respectively and $0$  the machine elements. Note that job $4$ represented by $x_4$ is not selected in the partial schedule.} 
    \centering
    \includegraphics[width=0.6\textwidth]{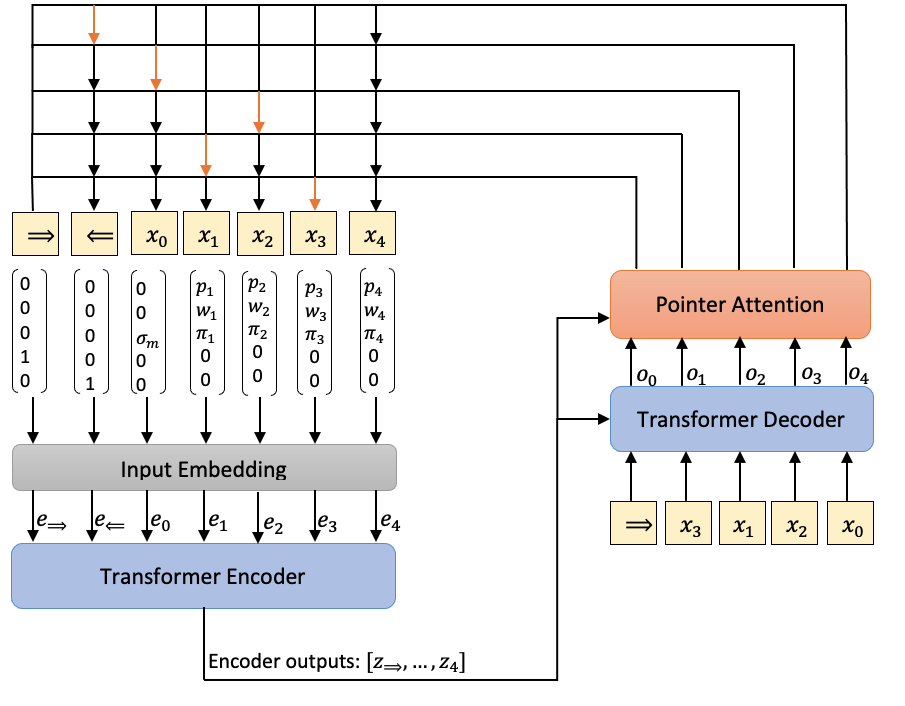}\label{fig:PointerTransformer}
\end{figure}

Figure \ref{fig:PointerTransformer} illustrates the proposed neural network architecture for an example single machine scheduling subproblem with 4 jobs. In this example, the input matrix $X \in R^{7 \times 5}$  involves 7 elements: $\Rightarrow$, $\Leftarrow$, machine, and four jobs, each of which is represented by a $1\times 5$  vector. We first pass $X$ through an embedding layer that projects each input vector to a  $d$ dimensional space, where $d$ is a hyperparameter (set to 64 through hyperparameter tuning described in Section~\ref{hyperparameter}). The matrix operation in the embedding layer is $E = XW_{\text{embed}}$, where $W_{\text{embed}} \in R^{5 \times d}$ is a weight matrix learned by training. Each row $e_i$ of $E$  represents the embedding corresponding to input $x_i$. The embedding layer facilitates the training of the model by providing smooth, continuous input to the network. The total number of learnable parameters in the embedding layer is given by the number of elements in $W_{\text{embed}} \in R^{5 \times d}$. In our case, with 5 input features and $d = 64$, $5\times 64 = 320$ weights are trained for this portion of the architecture.

\subsubsection{Transformer Network}
\begin{figure}[ht]
    \centering
    \caption{Transformer-based Encoder and Decoder. Figure adapted from ~\citet{vaswani2017attention} where $N$ is the number of layers.}
    \includegraphics[width=0.6\textwidth]{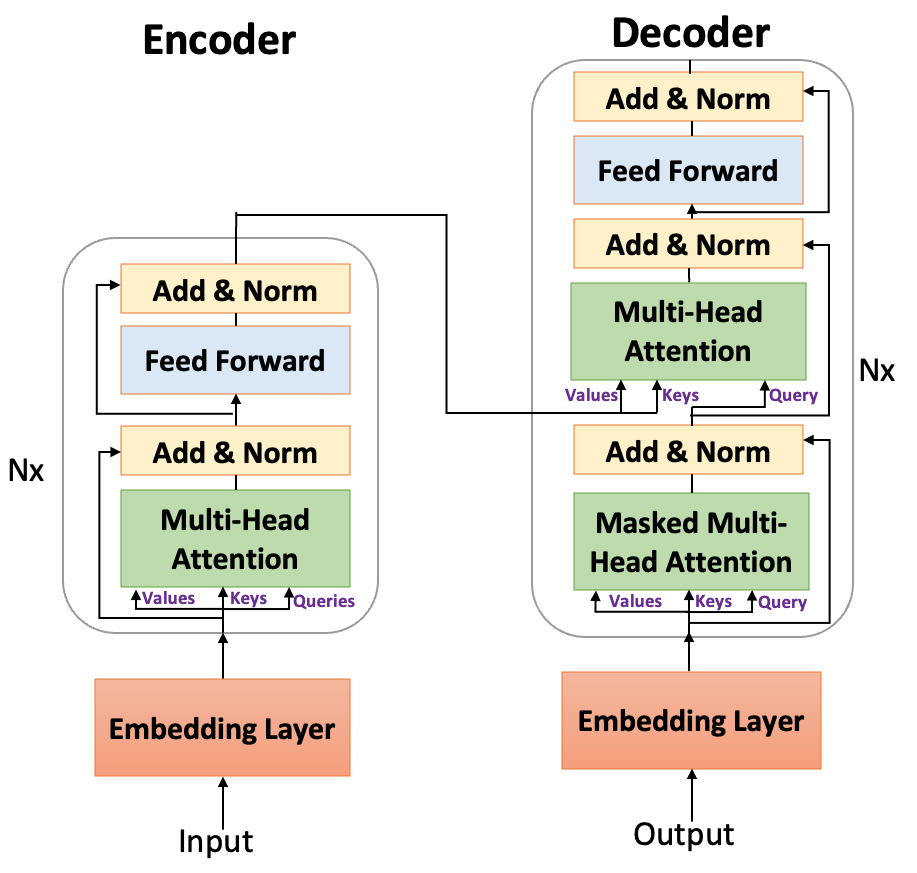}
    
    \label{fig:vaswani}
\end{figure}

We employ the transformer architecture ~\citep{vaswani2017attention} depicted in Figure~\ref{fig:vaswani} as a baseline. A transformer consists of an encoder and a decoder, both of which utilize the \textit{multi-head attention mechanism}, which plays a crucial role in capturing dependencies within an input sequence, essential for modeling  interactions between jobs and the machine in our pricing subproblem. 

\noindent \textbf{Attention Mechanism}. 
The attention mechanism was introduced to improve the performance of neural networks in tasks involving long sequences, such as machine translation. 
In its most general form, the attention mechanism computes a weighted sum of  \textit{values} corresponding to elements in the input sequence. The weights are determined by the similarity (often through dot-product or other scoring functions) between a set of \textit{queries} and \textit{keys}, where the queries search for the information that a model is looking for, and the keys represent various characteristics of the input elements. If keys, queries, and values are derived from the same input sequence, the mechanism is referred to as \textit{ self-attention} and allows tokens of the sequence to attend to each other. Mathematically, given an embedding matrix $E\in R^{n \times d}$ for an input sequence with $n$ tokens,  let $Q=EW^Q+\mathbf{1}_nb^Q$, $K=EW^K+\mathbf{1}_nb^K$ and $V=EW^V+\mathbf{1}_nb^V$ for column vector of ones $\mathbf{1}_n \in R^{n\times 1}$, weight matrices $W^Q \in R^{d \times d_k}$, $W^K \in R^{d \times d_k}$, $W^V \in R^{d \times d_v}$, and bias vectors $b^Q,b^K \in R^{1 \times d_k}$, $b^V \in R^{1 \times d_v}$. The resulting $Q \in R^{n\times d_k}, K \in R^{n\times d_k}$ and $V \in R^{n\times d_v}$ matrices are referred to as the \textit{query}, \textit{key}, and \textit{value} matrices, respectively. The attention score is then computed as:
\begin{equation}
       \text{Attention}(Q,K,V) =  \text{ softmax}(\frac{QK^T}{\sqrt{d_k}})V. \label{Attention}
\end{equation}
The softmax operation normalizes the weights in each column of the attention matrix between 0 and 1 in such a way that they add up to 1. 
The final matrix obtained from the attention mechanism is in $R^{n \times d_v}$ and captures the relation between each pair of tokens in the input sequence.\\
\noindent\textbf{Masked Attention}
During training, the entire target sequence is available, and all sequence positions are generated (decoded) in parallel.
Therefore, in the decoder part of the transformer, a masked version of self-attention is used to ensure that only  tokens decoded in the earlier steps $s_0, s_1, ..., s_{t-1}$ but not future tokens are considered when decoding the next token $s_t$. For a target schedule of length $n$, the mask takes the form of an $n \times n$ upper triangular matrix $M$ given by: 
\[
M_{ij} = 
\begin{cases}
    0 & \text{if } j \leq i \\
    -\infty & \text{if } j > i
\end{cases}
\]
The masked attention mechanism is then computed by:
\begin{equation}
       \text{Attention}(Q,K,V) =  \text{ softmax}(\frac{QK^T}{\sqrt{d_k}} + M)V. \label{MaskedAttention}
\end{equation}
 This mask fills the attention score for the positions of future tokens with negative infinity.  Masking is also required during inference because the model's self-attention is inherently parallel and operates over all positions in the sequence, even when future tokens have not yet been generated. The mask prevents the model from mistakenly trying to attend to future positions, ensuring that each token is generated autoregressively based only on previously generated tokens.\\
\noindent \textbf{Multi-Head Attention.} The transformer utilizes a multi-head attention mechanism by deriving multiple query, key, and value matrices, each with its own set of weight matrices. The self-attention is then applied to these representations in parallel. The outputs of the attention heads are concatenated and linearly projected to generate the final output. In particular,
\begin{equation}
 \text{MultiHead}(Q,K,V) =  \text{Concat(head}_1, ..., \text{head}_h)W^O + \mathbf{1}_nb^O
 \label{multihead}
\end{equation}
where $\text{head}_i = \text{Attention}(Q_i,K_i,V_i) \in R^{n \times d_v}$ for $i = 1, ..., h$, the outer projection matrix $W^O \in R^{hd_v \times d}$ and bias vector $b^O \in R^{1 \times d}$. We use $d_k$ = $d_v = \frac{d}{h}$ where the number of heads $h=8$ and hidden dimension $d=64$.
Thus, for each head $i$, we have $W^Q_i \in R^{64 \times 8} \text{ with } b^Q_i \in R^{1\times 8}$,$W^K_i \in R^{64 \times 8} \text{ with } b^K_i \in R^{1 \times 8}$, and $W^V \in R^{64 \times 8} \text{ with } b^V_i \in R^{1 \times 8}$
and there are $3\times (64\times 8 + 8) = 1,560$ learnable weights per head. Furthermore, the outer projection matrix $W^O \in R^{64 \times 64} \text{ with 
 } b^O \in R^{1 \times 64}$. Thus, the total number of learnable parameters for the multi-head attention mechanism with 8 heads is $8 \times 1,560 + 64 \times 64 + 64 = 16,640$.
 
\noindent \textbf{Transformer Encoder.}
The transformer encoder stacks $N$ sequential layers, each of which comprises two sub-layers: a multi-head self-attention sublayer and a fully connected feed-forward  network, as shown in Figure~\ref{fig:vaswani}. The output of the multi-head self-attention is passed through a feed-forward network. In the first encoder layer, $Q, K, V$ matrices are derived from the input embedding $E$. In the subsequent encoder layers, these matrices are derived from the output of the previous layer.  The output of the encoder is denoted by a context matrix $Z\in R^{n\times d}$. Each row of the context matrix corresponds to a specific token of the input sequence and captures the interactions between that token and all others in the same sequence. 
\sloppy
Layer normalization is applied after each sublayer through $\texttt{LayerNorm}(g + \texttt{Sublayer}(g))$, where $\texttt{Sublayer}(g) \in R^d$ is the output of the sublayer (e.g. multi-head self-attention) and $g \in R^d$ is the input to the sublayer (e.g. embedding for each input token). Let $\varrho = g + \texttt{Sublayer}(g)$, $\mu = \sum_{i=1}^d\varrho_i/d$ and $\sigma= \sqrt{\frac{1}{d}\sum_{i=1}^d(\varrho_i-\mu)^2}$ then 
$$\texttt{LayerNorm}(\varrho) = \frac{\varrho-\mu}{\sigma}\odot w_L + b_L,$$
where $\odot$ represents element-wise multiplication, and $w_L\in R^d$ and $b_L \in R^d$ are learnable normalization weight and bias vectors. Thus, for $d=64$,  each layer normalization has a total of $128$ learnable parameters.  In the transformer architecture, layer normalization helps reduce internal covariate shift and ensures that the input distribution remains consistent as it passes through the layers, enabling more stable training and faster convergence.  Since layer normalization operates independently on each token, it is well-suited for sequence models where input sizes or sequences may vary~\citep{ba2016layer}.

The feed-forward sublayer consists of two linear layers, each with a $d \times d$ weight matrix 
 and a bias vector $\in R^{d}$. Thus, for $d=64$, the total number of learnable parameters in the feed-forward sublayer is $2 \times (64 \times 64 + 64) = 8,320$. The number of learnable parameters per encoder layer is the total number of weights for the multi-head self-attention sublayer, feed-forward sublayer, and two layer normalization which is: $16,640 + 8,320 + 2 \times 128 = 25,216$. For $N=2$ encoder layers, the total number of encoder parameters is $2 \times 25,216 = 50,432$.

\noindent \textbf{Transformer Decoder.}
The decoder in the transformer has a similar structure to the encoder, consisting of three stacked layers, each in turn consisting of three sublayers. A masked multi-head self-attention sublayer is applied at the beginning, a feed-forward sublayer at the end of each layer and another multi-head \textit{cross-attention} sublayer in the middle (Figure~\ref{fig:vaswani}).  

\begin{figure}[htbp]
 \caption{Autoregressive Inference of the Decoder. In the first step, the decoder takes as input the $\Rightarrow$ and the encoder output matrix $Z$, producing $o_0$. The pointer layer then uses $o_0$ and $Z$ to select $x_3$, as shown by the orange arrow. In the subsequent steps, the previously generated tokens are passed to the decoder and excluded from further selection in the pointer attention layer.\\}
    \centering
    
    \begin{subfigure}[b]{0.5\textwidth}
        \includegraphics[width=\textwidth]{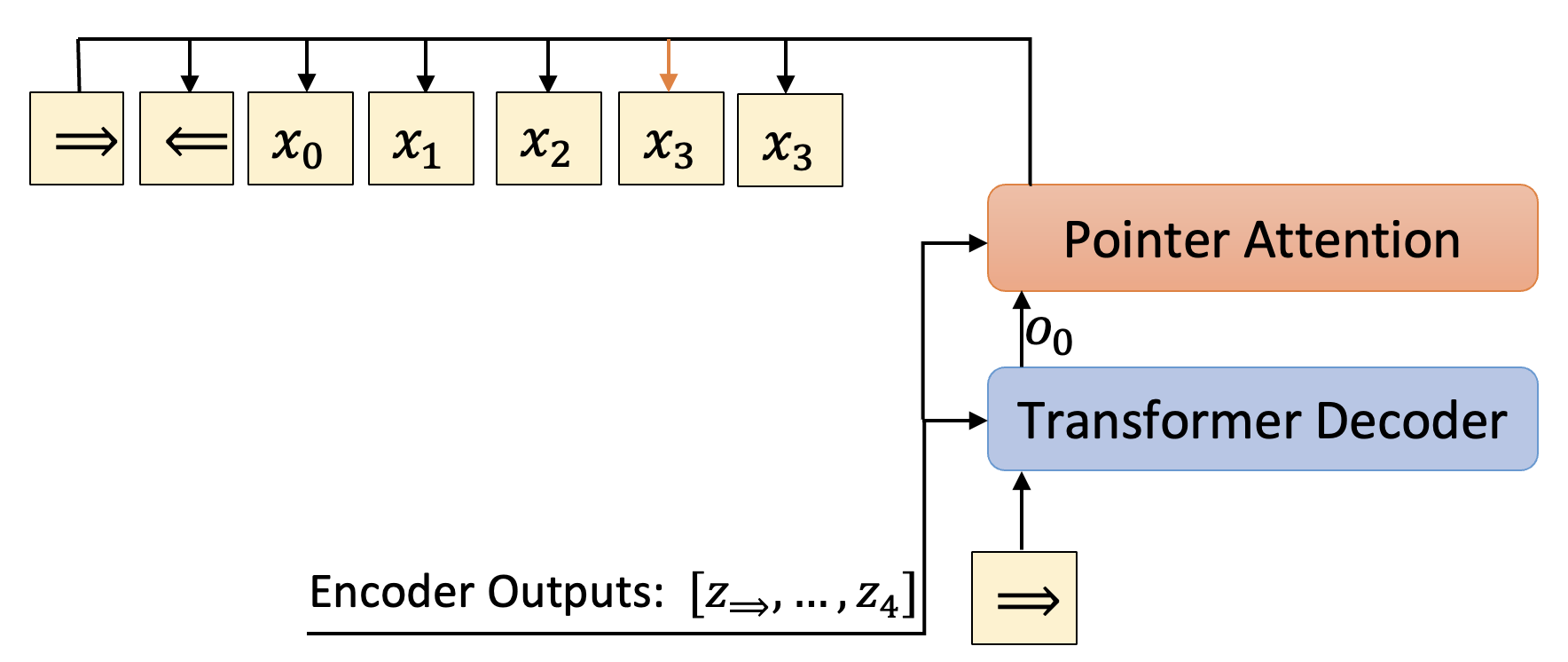} \caption{First Decoding Step}
        \label{fig:sub2}
    \end{subfigure}
    \begin{subfigure}[b]{0.5\textwidth}
        \includegraphics[width=\textwidth]{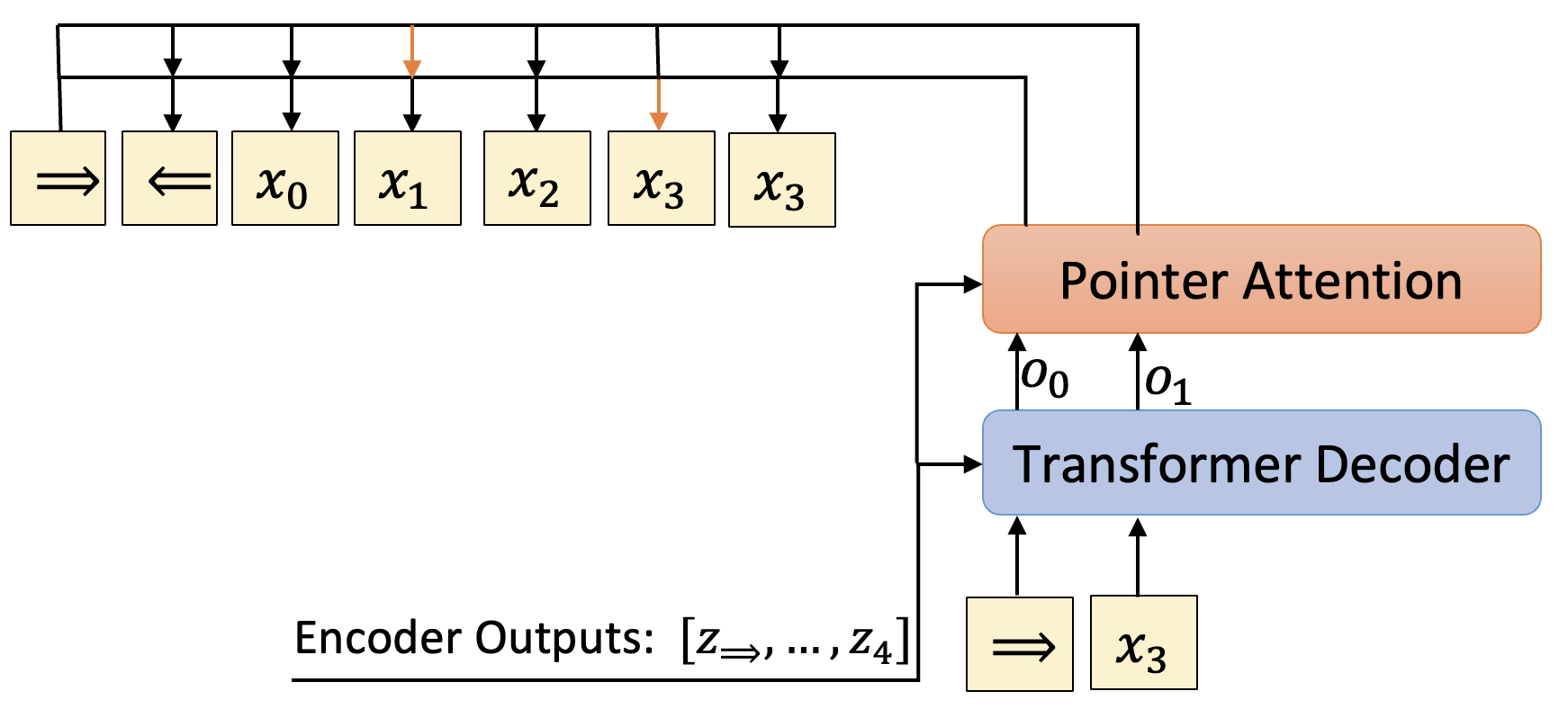} \caption{Second Decoding Step}
        \label{fig:sub3}
    \end{subfigure}
    \begin{subfigure}[b]{0.5\textwidth}
             \includegraphics[width=\textwidth]{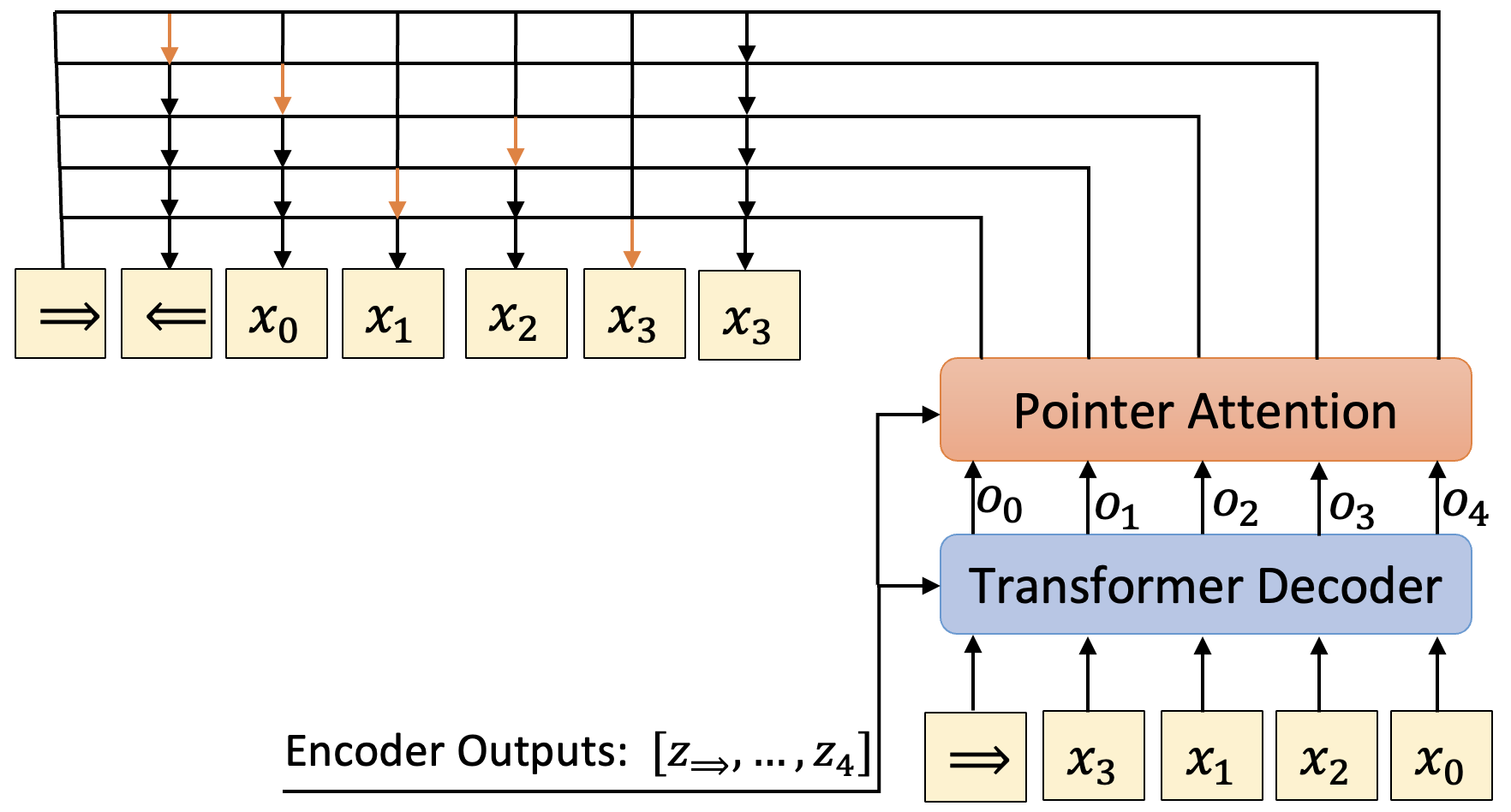}\caption{Last Decoding Step}
        \label{fig:sub4}
    \end{subfigure}
    \label{fig:multi}
\end{figure}

The inference process in the transformer decoder is auto-regressive, the model generates one token at a time and feeds it back as input for generating the next token, i.e. schedule position. As shown in Figure~\ref{fig:multi}, in the first decoding step, the input to the decoder is the encoder output $Z$ along with $\Rightarrow$ indicating the beginning of a schedule.  In the subsequent decoding steps, the decoder takes as input the output of the encoder along with the previously generated elements $(s_{0}, s_{1}, ..., s_{i-1})$ to create an output  $o_i \in R^{d}$, encapsulating knowledge of the previously generated schedule elements up to the current decoding step and the contextual information provided by the encoder. 

In the first sub-layer of the decoder, the masked multi-head attention uses the embedding of the last generated token $s_{i-1}$ as the query, while the keys and values are the embeddings of the previously generated tokens $(s_{0}, s_{1}, ..., s_{i-1})$. The mask ensures that the model does not attend to any tokens beyond the current position, preserving the autoregressive nature of the transformer. In the subsequent cross-attention sub-layer, the query is the output of the masked multi-head attention layer, while the keys and values are the encoder output $Z$. This allows the decoder to use the information from the input sequence while generating the output.

 The total number of parameters per decoder layer is given by the number of parameters of the two multi-head self-attention sublayers ($2 \times 16,640$), feed-forward sublayer ($8,320$), and three layer normalization ($3 \times 128$), which equals $41,984$. For $N=2$ decoder layers, the total number of parameters is equal to $2 \times 41,984 = 83,968$.
In a traditional transformer network, the decoder output $o_i$ is passed through a linear transformation followed by a softmax to generate a probability distribution over the target output space. In this study, however, we pass $o_i$ to a \textit{pointer attention} layer, as detailed below.

\subsubsection{Pointer Attention Layer}
\label{pointer sec}
We obtained only a 45\% validation accuracy using the traditional transformer.
To address this challenge, we replaced the final linear layer in the transformer architecture with a pointer layer. The pointer layer uses attention to selectively point back to elements in the input set at each decoding step, allowing the model to dynamically focus on the most relevant elements. This is accomplished by combining the encoder output matrix $Z$ and the decoder output vector $o_i$ at decoding step $i$ into attention weights $u_j^i$  that represent the importance of each element $j$ in the input with respect to the current decoding step $o_i$. The pointer attention score $u_j^i$ is computed as:
\begin{equation*}
    u_j^i = \begin{cases}
    v^T\tanh(W_1z_j^T + W_2o_i^T) & \text{if } j \notin \{s_0,\ldots,s_{i-1}\}  \\
    -\infty & \text{otherwise,}
    \end{cases} \label{pointer}
\end{equation*}
where $v \in \mathbb{R}^d$, $W_1 \in \mathbb{R}^{d \times d}$, and $W_2 \in \mathbb{R}^{d \times d}$ are learnable parameters of the model, represented as linear layers. The $\tanh$ activation function captures the non-linear relationship between the encoder output $z_j$ for input token $j$ and the decoder output $o_i$. To obtain the probability distribution $p(s_i|s_1, \ldots, s_{i-1},X)$, indicating the likelihood of selecting element $s_i$ given the previous selections $s_1, \ldots, s_{i-1}$ and the input $X$, the softmax function is applied to the attention vector $u^i$, i.e. $p(s_i|s_1, . . . , s_{i-1},X) = \text{softmax}(u^i)$.

The softmax function normalizes the vector $u^i$ to create an output distribution over the input set. By setting the logits (log-probabilities) of input elements that have already appeared in the schedule to $-\infty$, we guarantee that the model only focuses on elements that have not yet been generated, preventing duplicate selections. The pointer network learnable matrices are $W_1 \in R^{64 \times 64}, W_2 \in R^{64 \times 64}$, and vector $v \in R^{64}$. Thus, the number of learnable parameters for the pointer network is
 $2 \times 64 \times 64 + 64 =  8,256$.
The total learnable parameters in the transformer-pointer architecture is given by the number of embedding, encoder, decoder, and pointer weights, for a total of $320 + 50,432 + 83,968 + 8,256 = 142,976$ learnable parameters.
\section{Data Generation}
\label{Data Generation}
To generate the dataset for training and validation, the column generation algorithm is executed by solving the pricing subproblems using dynamic programming. We generate instances with different number of jobs $N \in \{20, 30, 40, 60, 80\}$ and machines $M \in \{2, 4, 8, 12, 16\}$. In the first set of experiments, the job weights $w_j$ and processing times $p_{jk}$ are generated following~\citet{chen1999solving}, with integer weights uniformly drawn from $[1,100]$ and integer processing times uniformly sampled from $[1,30]$. 

The testing data include instances generated using the same uniform distributions as well as out-of-distribtion (OOD) instances in which the weights are independent integers drawn from a Weibull distribution with shape parameter $k=1.5$ and scale parameter $\lambda=50$, while the processing times are integers independently sampled from a Weibull distribution with shape parameter $k=2$ and scale parameter $\lambda=15$. We refer to the combination of the number of machines and the number of jobs as a problem class (for example, $2$ machines and $20$ jobs) and generate multiple instances from each class. We start the column generation with a specific number (5, 10, 20, 50, or 100) of initial columns per machine.

In each iteration of the column generation, the processing times and weights remain constant, but we obtain new dual values for jobs and machines. Using these updated dual values, we solve each subproblem using dynamic programming to generate a new negative reduced cost column per machine which serves as the target output $s$ of the neural network. If a subproblem finds no negative reduced cost column, it returns an empty vector. Hence, after each CG iteration, we obtain one new data instance per machine.

We generate $73,144$ data instances, of which $80\%$ are used for training and 20\% for validation. Table~\ref{CG Problem Classes and the Associated  Generated NN Instances} in Appendix~\ref{CG Problem Instances appendix} lists the percentage of training instances (data points) per problem class. Most problem classes have similar percentages except for those involving 2 machines and 20 jobs, and those with 4 machines and 60 jobs, which have the highest percentage of data points. The difference arises because it is difficult to control the exact number of subproblems that will be solved in the course of a given CG run. Although the data are not evenly distributed across the different problem classes, our results demonstrate that our model can effectively generalize to problems of varying sizes.

\section{Computational Experiments}
\label{experiments}
This section provides an overview of the training procedure, hyperparameter tuning, and the comparison of testing instances with the traditional column generation approach employing the dynamic programming procedure with the greedy selection of the most negative reduced-cost column. The entire implementation is done in Python. The restricted master problem is solved using Gurobi 9.1.2,  while the training and testing of the neural network are conducted using PyTorch 1.13.1. We run the hyperparameter tuning and training on two Nvidia A100 GPUs. The testing and data generation are executed on an Apple M2 Pro.

\subsection{Training}
During the training process, we employ the teacher-forcing strategy ~\citep{williams1989learning}, which involves passing the previous steps of the target job sequence to the next decoding step, rather than using the model's predicted output. For instance, in the second decoding step in Figure~\ref{fig:multi}, the true target job in the first position is $x_3$. Therefore,  the input to the decoder in the second decoding step is $\Rightarrow$ and $x_3$ regardless of the model's prediction for the first position, which might not be $x_3$.

We train the model using a negative log-likelihood loss function $l(\theta)$ measuring the cross-entropy between the predicted probabilities and the target schedule generated by the DP. Specifically, 
\begin{center}
  $l(\theta) = - \sum_{i=1}^n \log p(s_{i}|X; \theta),$
\end{center}
where $s_i$ is the target element in schedule position $i$.
We employ the Adam optimizer~\citep{kingman2015adam} with  parameter values found through hyperparameter tuning. 
We assess the prediction performance using accuracy, which is calculated as the percentage of correctly predicted sequence positions.

\subsection{Hyperparameter Tuning}
\label{hyperparameter}
Hyperparameter tuning is a crucial step in optimizing the performance of a neural network model. Table~\ref{HyperparameterValues} includes the list of parameters along with their tested and the best-found  values. The tested values for each parameter are chosen from the literature, and descriptions of these hyperparameters are given in Appendix~\ref{Hyperparameter tuning appendix}.
To find the best configuration of these hyperparameters, we use RayTune~\citep{liaw2018tune}, a Python library that offers state-of-the-art tuning algorithms with the advantage of parallel computing. We use the Asynchronous Successive Halving Algorithm~\citep{li2020system}, which combines random search with principled early stopping. This algorithm samples different combinations of hyperparameter values given in Table~\ref{HyperparameterValues}, and trains models in parallel. We conduct 100 trials, that is, sample 100 hyper parameter configurations (see Table~\ref{Evaluated HyperParameter Tuning Configurations and Validation performances} in Appendix \ref{Hyperparameter tuning appendix}). Each trial involves training the model for 150 epochs. A complete pass through the training dataset is made during each epoch. The validation accuracy is used to evaluate the prediction performance, and a grace period of 5 epochs is set for early stopping. If the accuracy does not improve within the grace period, the trial is terminated early.

The results of the hyperparameter tuning, including the validation loss and accuracy over 150 training epochs, are presented in Figures \ref{fig:LossEpochRayTune} and \ref{fig:AccuracyEpochRayTune}, respectively, in Appendix \ref{Hyperparameter tuning appendix}. Early stopped trials are represented by lines that terminate before reaching the maximum number of epochs. These trials exhibit a plateau in the validation loss, indicating that further training is unlikely to yield better results. After analyzing the initial 100 trials, we conducted an additional round of hyperparameter tuning that explored various key hyperparameters, including larger model dimensions (128 and 256), reduced weight decay ($1e{-07}$), an increased number of heads (16), larger batch sizes (64 and 128), and lower dropout rates. However, none outperformed the best configuration from the initial tuning.  

\begin{table}[htbp]
    \centering
    \caption{Tested Values of the Hyperparameters and the Best Configuration}
    \begin{tabular}{lll}
    \hline
        Hyperparameter & Tested Values & Best Found \\ \hline
        Model Dimension & 16, 32, 64 & 64 \\ 
        Number of Attention Heads & 2, 4, 8 & 8 \\ 
        Number of Encoder Layers  & 1, 2, 3, 6 & 2\\ 
        Number of  Decoder Layers  & 1, 2, 3, 6 & 2\\ 
        Dropout & 0.0, 0.3, 0.5, 0.8 & 0.0\\ 
        Weight Decay & 0, 0.1, 0.01, 1e-3, 1e-5, 1e-6 & 1e-6\\       
        Batch Size & 16, 32 & 16\\ 
        Learning Rate & 1e-5, 1e-4, 1e-3, 1e-2 & 1e-4\\ 
        \hline
    \end{tabular}
    \label{HyperparameterValues}
\end{table}
\section{Results}
\label{Results}
\subsection{Training and Validation Results}
\begin{figure}[htbp]
    \caption{Training and Validation Loss and Accuracy}
    \centering
        \begin{subfigure}[b]{0.7\textwidth}
        \includegraphics[width=\textwidth]{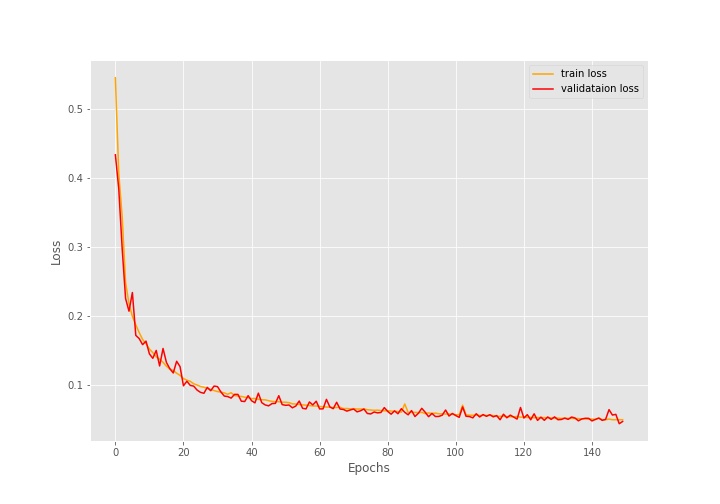} \caption{Training and validation loss}
    \end{subfigure}
        \begin{subfigure}[b]{0.7\textwidth}
        \includegraphics[width=\textwidth]{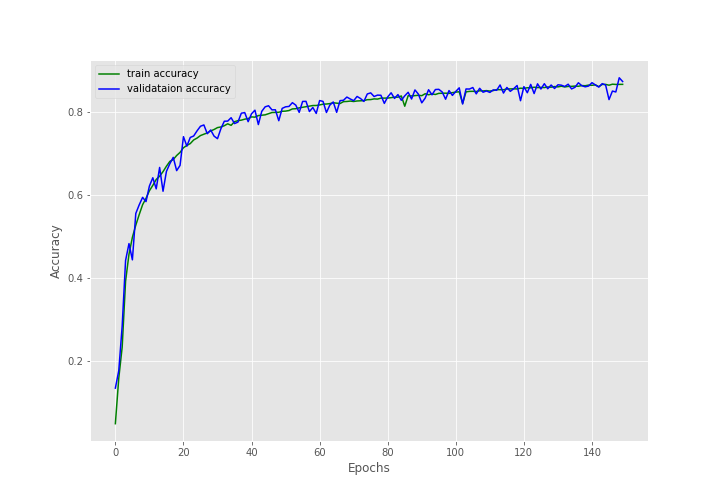} \caption{Training and validation accuracy}
    \end{subfigure}

    \label{fig:modellossacc}
\end{figure}

The training and validation loss and accuracy for the best-found model are shown in Figure ~\ref{fig:modellossacc}. The training loss demonstrates a decreasing trend, indicating that the model is effectively learning. The validation loss curve shows a decreasing trend, indicating the model's ability to generalize to unseen data. The fact that the validation loss is similar to the training loss suggests that the model is not overfitting. The validation accuracy reaches  $86\%$, indicating that on average the model correctly predicts $86\%$ of the positions in the target schedules in the unseen validation data. 

\subsection{Testing Results}

We conduct experiments to investigate the performance of the proposed method compared to that using dynamic programming to solve the pricing subproblems. We apply the SWPT order rule to the NN  schedules minimizing the total completion time on the machine. The first set of experiments involves small to medium size problem instances, as shown in Table~\ref{Per Instance Optimal Results}. Instance name (e.g. 2M20N\_1\_20) indicates the number of machines (2), number of jobs (20), seed number (1), and the number of initial columns per machine (20). We solve these instances to optimality using the baseline (traditional column generation (CG) with greedy dynamic programming) and the proposed approach.  The table shows the total time required to reach optimality and the total number of generated columns for both approaches. The proposed approach achieves a significant time reduction as the instance size increases and generally adds more columns than the baseline approach. 
Interestingly, \texttt{CG NN-DP} produces fewer columns than \texttt{CG-Greedy-DP} in some smaller instances, such as 4M20N\_1\_10, 4M20N\_3\_10, 4M20N\_4\_10, and 4M30N\_2\_10; number of columns is shown in red in the table when NN generates fewer columns than DP. 

\begin{table}[!ht]
\small
    \caption{Computational Results for Small and Medium Size Instances}
    \begin{tabular}{c|ccc|c|ccc}
    \hline
        \textbf{} & \multicolumn{3}{|c|}{Total Time (sec)} &  \multicolumn{4}{|c} {Number of Columns Generated} \\ \hline
        \textbf{ } &   &  & Time  & \multicolumn{1}{c}{}   & \multicolumn{3}{c}{CG NN-DP} \\ \cline{6-8}
        Instance & CG-Greedy-DP & CG NN-DP & Reduction & \multicolumn{1}{c}{CG-Greedy-DP}  & \multicolumn{1}{c}{Total} & NN & DP \\ \hline
        2M20N\_1\_20& 45 & 30 & 33\% & 217 & 235 & 129 & 106 \\ 
        2M20N\_2\_20 & 39 & 20 & 49\% & 182 & \textcolor{red}{181} & 112 & 69 \\ 
        2M20N\_3\_20 & 34 & 29 & 14\% & 193 & 244 & 133 & 111 \\
        2M20N\_4\_20 & 31 & 13 & 58\% & 175 & \textcolor{red}{170} & 135 & 35 \\ 
        2M20N\_5\_20 & 39 & 26 & 33\% & 204 & 241 & 144 & 97 \\ 
        2M20N\_6\_10 & 35 & 20 & 43\% & 200 & 237 & 162 & 75 \\
        2M20N\_7\_10 & 32 & 18 & 42\% & 194 &  206 & 133 & 73 \\ 
        2M20N\_8\_10 & 36 & 20 & 44\% & 215 & \textcolor{red}{210} & 125 & 85 \\ 
        2M20N\_9\_10 & 46 & 35 & 24\% & 214 & 219 & 99 & 120 \\ 
        2M20N\_10\_10 & 44 & 24 & 45\% & 212 &  242 & 151 & 91 \\ 
        \hline
        \textbf{Average} & \textbf{38.1} & \textbf{23.5} & \textbf{38.5\%}& \textbf{200.6} & \textbf{218.5} & \textbf{132.2} & \textbf{86.2}\\
        \hline
        4M20N\_1\_10 & 45 & 30 & 33\% & 215 & \textcolor{red}{180} & 84 & 96 \\ 
        4M20N\_2\_10 & 35 & 20 & 43\% & 184 & 185 & 115 & 70 \\ 
        4M20N\_3\_10 & 31 & 20 & 35\% & 176 & \textcolor{red}{172} & 97 & 75  \\ 
        4M20N\_4\_10& 37 & 11 & 70\% & 192 & \textcolor{red}{159} & 125 & 34 \\ 
        4M20N\_5\_10 & 42 & 22 & 48\% & 215 & \textcolor{red}{209} & 153 & 56 \\ 
        4M20N\_6\_5 & 36 & 22 & 37\% & 198 & 200 & 121 & 79 \\
        4M20N\_7\_5& 40 & 19 & 52\% & 218 & 218 & 158 & 60 \\ 
        4M20N\_8\_5 & 38 & 17 & 55\% & 222 & \textcolor{red}{217} & 153 & 64 \\ 
        4M20N\_9\_5 & 53 & 29 & 45\% & 256 & \textcolor{red}{231} & 123 & 108 \\ 
        4M20N\_10\_5& 45 & 25 & 44\% & 217 & \textcolor{red}{202} & 109 & 93 \\
        \hline
        \textbf{Average} & \textbf{40.2} & \textbf{21.5} & \textbf{46.2\%}& \textbf{209.3} & \textcolor{red}{\textbf{197.3}} & \textbf{123.8} & \textbf{73.5}\\
        \hline
        2M30N\_1\_20 & 309 & 151 & 51\% & 462 &  551 & 376 & 175 \\ 
        2M30N\_2\_20 & 336 & 96 & 71\% & 501 & 556 & 444 & 112 \\ 
        2M30N\_3\_20 & 267 & 153 & 43\% & 499 & 631 & 417 & 214 \\ 
        2M30N\_4\_20 & 248 & 152 & 39\% & 420 &  598 & 391  & 207 \\ 
        2M30N\_5\_20 & 347 & 139 & 60\% & 575 & 590 & 408 & 182 \\ 
        2M30N\_6\_10 & 270 & 164 & 39\% & 466 &  632 & 387 & 245 \\ 
        2M30N\_7\_10 & 327 & 143 & 56\% & 558 & 623 & 434 & 189 \\ 
        2M30N\_8\_10 & 242 & 121 & 50\% & 495 & 558 & 369 & 189 \\ 
        2M30N\_9\_10 & 336 &  165 & 51\% & 526 & 622 & 404 & 218 \\ 
        2M30N\_10\_10 & 308 & 170 & 45\% & 486 &  625 & 386 &  239\\ 
        \hline
        \textbf{Average} & \textbf{299} & \textbf{145.4} & \textbf{50.5\%}& \textbf{498.8} & \textbf{598.6} & \textbf{401.6} & \textbf{197}\\
        \hline
        4M30N\_1\_10 & 294 & 157 & 46\% & 457 & 466 & 269 & 197 \\ 
        4M30N\_2\_10 & 355 & 173 & 51\% & 511 & \textcolor{red}{455} & 321 & 134 \\ 
        4M30N\_3\_10 & 237 & 87 & 63\% & 430 & \textcolor{red}{392} & 295 & 97 \\ 
        4M30N\_4\_10 & 298 & 155 & 48\% & 491 & 512 & 361 & 151 \\ 
        4M30N\_5\_10 & 330 & 172 & 47\% & 507 & \textcolor{red}{503} & 311 & 192 \\ 
        4M30N\_6\_5 & 265 & 123 & 53\% & 476 & \textcolor{red}{470} & 332 & 138 \\ 
        4M30N\_7\_5 & 237 & 135 & 43\% & 421 & 512 & 351 & 161 \\
        4M30N\_8\_5 & 266 & 80 & 70\% & 498 & \textcolor{red}{439} & 325  & 114 \\ 
        4M30N\_9\_5 & 303 & 126 & 58\% & 514 & 528 & 357 & 171 \\ 
        4M30N\_10\_5 & 352 & 94 & 73\% & 517 & \textcolor{red}{461} & 355 & 106 \\
        \hline
        \textbf{Average} & \textbf{287.2} & \textbf{134.2} & \textbf{53.2\%}& \textbf{478.3} & \textcolor{red}{\textbf{475.2}} & \textbf{324.7} & \textbf{150.6}\\
        \hline
        2M40N\_1\_20 & 1929 & 1140 & 41\% & 1117 & 1364 & 639 & 725  \\ 
        \hline
    \end{tabular}
    \label{Per Instance Optimal Results}
\end{table}

\begin{figure}
\centering
\caption{Convergence plots of different CG-DP approaches and \texttt{CG-NN-DP} for test instances 8M60N, 16M80N, and 20M100N generated form Uniform distribution. The solid curves are the mean of the relative objective values over 10 instances and the shaded area shows ±1 standard deviation.} \label{fig:8m60n20m100n}
    \begin{subfigure}[b]{0.6\textwidth}
        \includegraphics[width=\textwidth]{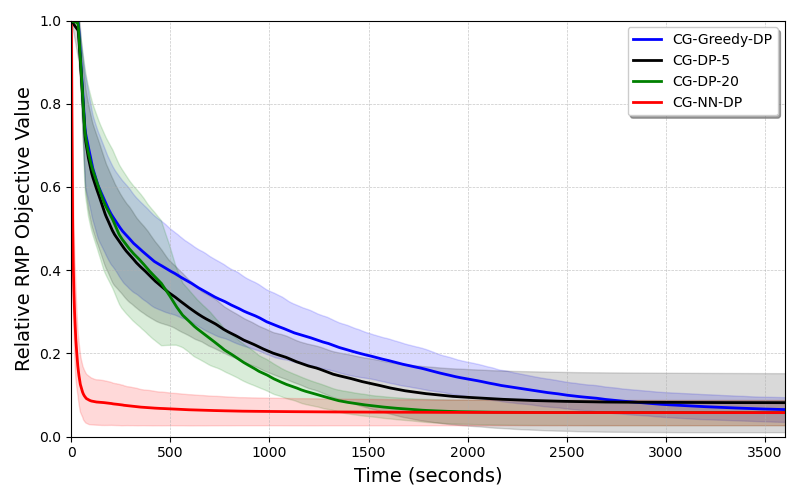} 
        \caption{8 Machines 60 Jobs (8M60N)} \label{fig:8M60N}
    \end{subfigure}
    \begin{subfigure}[b]{0.6\textwidth}
        \includegraphics[width=\textwidth]{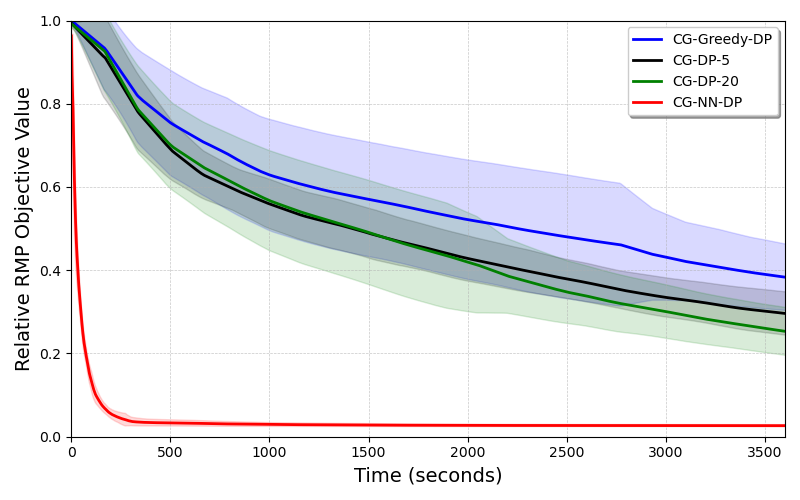} 
        \caption{16 Machines 80 Jobs (16M80N)} \label{fig:16M80N}
    \end{subfigure}
    \begin{subfigure}[b]{0.6\textwidth}
        \includegraphics[width=\textwidth]{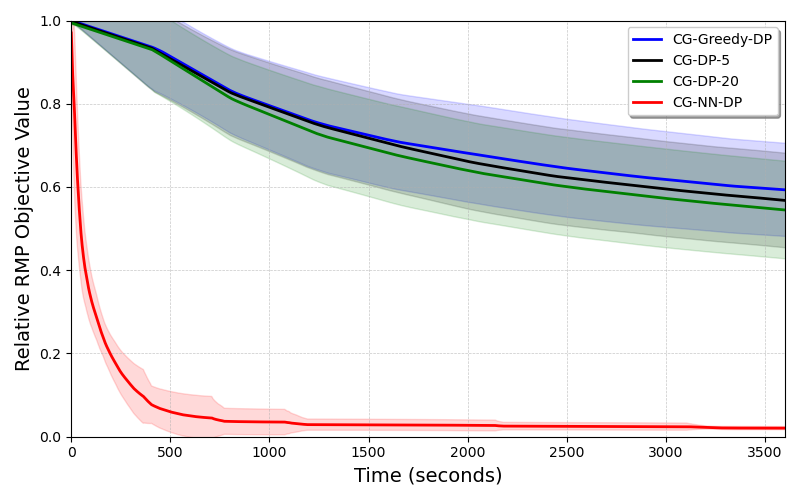} 
        \caption{20 Machines 100 Jobs (20M100N)} \label{fig:20M100N}
    \end{subfigure}
\end{figure}

To show the performance of the proposed approach on  large-scale instances, we compare it to multiple versions of column generation with DP. Specifically, we implement the \texttt{CG-Greedy-DP}, \texttt{CG-DP-5}, which adds the 5 most negative reduced columns per subproblem at each iteration, and \texttt{CG-DP-20}, which adds the 20 most negative reduced columns per subproblem at each iteration. Figure~\ref{fig:8m60n20m100n} shows the convergence plots for large-scale test instances with  8 machine 60 jobs, 16 machines 80 jobs, and 20 machines 100 jobs. We run 10 instances from each size for one hour, record the objective values of the RMP at each CG iteration and normalize them to be between $[0,1]$ before taking the average over all instances of a given problem class. \texttt{CG-NN-DP} consistently outperforms other methods. Furthermore, as seen in Figure~\ref{fig:8M60N}, \texttt{CG-NN-DP}  significantly improves the objective value within 100 seconds.  As the instance size grows,  the \texttt{CG-DP} approaches struggle to improve the objective value even after 3,600 seconds compared to a rapid improvement by \texttt{CG-NN-DP} as shown in the convergence plot of  test instances with 20 machines and 100 jobs even though  \texttt{CG-NN-DP} adds a single column per subproblem at each iteration.Figure~\ref{fig:20M100N} shows that \texttt{CG-NN-DP}  performs very well for instances with 20 machines and 100 jobs, beyond the maximum training size of 80 jobs, demonstrating the generalizability of the proposed approach.

\begin{figure}
\centering
\caption{Convergence plots of different CG-DP approaches and \texttt{CG-NN-DP} for test instances 8M60N, 16M80N, and 20M100N generated from Weibull distribution. The solid curves are the mean of the relative objective values over 10 instances and the shaded area shows ±1 standard deviation.} \label{fig:8m60n20m100nWeibull}
    \begin{subfigure}[b]{0.6\textwidth}
        \includegraphics[width=\textwidth]{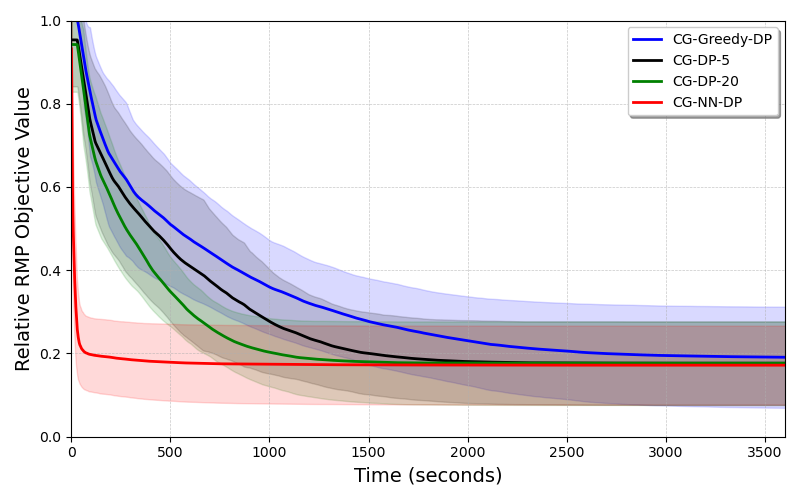} 
        \caption{8 Machines 60 Jobs (8M60N)} \label{fig:8M60NWeibull}
    \end{subfigure}
    \begin{subfigure}[b]{0.6\textwidth}
        \includegraphics[width=\textwidth]{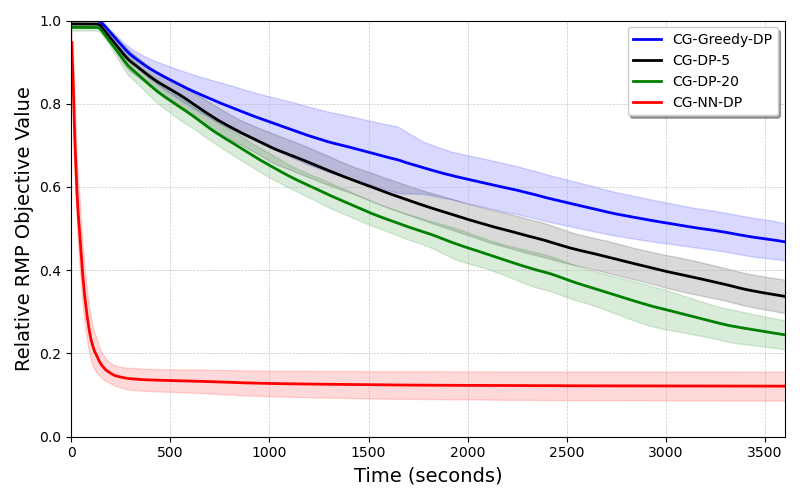} 
        \caption{16 Machines 80 Jobs (16M80N)} \label{fig:16M80NWeibull}
    \end{subfigure}
    \begin{subfigure}[b]{0.6\textwidth}
        \includegraphics[width=\textwidth]{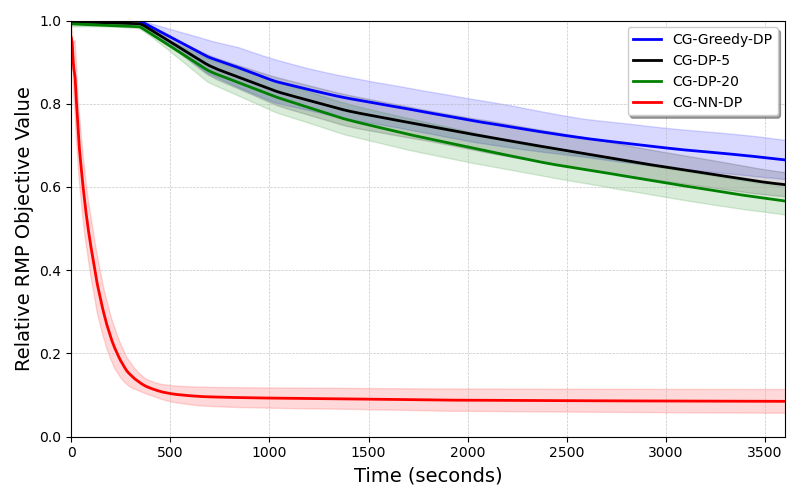} 
        \caption{20 Machines 100 Jobs (20M100N)} \label{fig:20M100NWeibull}
    \end{subfigure}
\end{figure}

We also test the trained neural network on instances generated from Weibull distribution to evaluate the generalizability of the \texttt{CG-NN-DP} method to a class of instances different from the training set consisting of instances generated from Uniform distribution. The results in Figure~\ref{fig:8m60n20m100nWeibull} again demonstrates the superior performance of \texttt{CG-NN-DP} compared to the other CG-DP approaches considered. 

\section{Conclusion and Future Research Directions}
\label{Conclusion}
In this paper, we train a transformer pointer network to rapidly approximate the dynamic programming solution of the pricing subproblem for minimizing the total weighted completion time on unrelated parallel machines. Our approach achieves significantly faster convergence than the traditional CG with dynamic programming, and exhibits generalization capability beyond the maximum number of jobs considered during training as well as to out-of-distribution instances. 

An interesting future direction is exploring the application of transfer learning~\citep{torrey2010transfer} to extend the proposed NN architecture to other scheduling problems. By leveraging the knowledge gained from solving the total weighted completion time problem on unrelated parallel machines, transfer learning could enable faster adaptation of the proposed neural network architecture to parallel machine scheduling problems with different performance measures. This approach has the potential to significantly reduce the need for extensive retraining on new problem instances, improving efficiency across a broader range of scheduling problems.

In this work, the NN adds a single column per machine at each iteration of the column generation process. As a future direction, we plan to explore the possibility of extending the approach to add multiple columns using beam search decoding~\citep{vijayakumar2016diverse} instead of greedy decoding. Beam search can generate multiple schedules (job sequences) by sampling various candidate jobs at each decoding step, rather than selecting only the schedule with the highest likelihood. By introducing this enhancement, we anticipate further improvements in the efficiency and effectiveness of the proposed \texttt{CG-NN-DP} approach.

Our supervised learning approach to training the transformer pointer network is susceptible to exposure bias~\citep{bengio2015scheduled}. To address this issue, we intend to investigate the application of reinforcement learning and self-supervised learning as alternative training methods. Both of these approaches would allow the neural network to learn based on the reduced costs of generated columns without relying on labeled data, but they follow distinct strategies. Reinforcement learning involves training the network in an environment and requires extensive exploration and the optimization of a reward function~\citep{wiering2012reinforcement}. In contrast, self-supervised learning requires large amounts of unlabeled data to learn inherent patterns in the problem structure~\citep{cai2024self}. 

Finally, it is worth considering the direct application of neural networks to the original parallel machine scheduling problem without relying on column generation. A neural network-based optimization proxy could be trained to approximate the solution to the original scheduling problem, potentially eliminating the need for the CG framework. This approach, however, might be more data-intensive and computationally demanding, especially for large instances.

\section{Acknowledgements}
\label{Acknowledgements}
This research was supported by the National Science Foundation (NSF)
under Grant CMMI-1826125. Any opinions
stated are those of the authors, and do not necessarily reflect those
of NSF.
\newpage
\bibliographystyle{trc}
\bibliography{ref}
\appendix 
\newpage
\section{Data points generated from each problem class}
\label{CG Problem Instances appendix}
\begin{table}[ht]
    \small
    \centering
    \caption{Data points generated from each problem class.}
    \begin{tabular}{cccc}
    \hline
        \textbf{Machines (M)} & \textbf{Jobs (N)} & \textbf{Percentage of data points} & \textbf{Number of data points} \\ \hline
        2 & 20 & 17.20\% & 12,576 \\ 
        2 & 30 & 2.30\% & 1,660 \\ 
        2 & 40 & 5.90\% & 4,308 \\ 
        4 & 20 & 3.50\% & 2,556 \\ 
        4 & 30 & 1.90\% & 1,384 \\ 
        4 & 40 & 5.90\% & 4,296 \\ 
        4 & 60 & 13.20\% & 9,664 \\ 
        8 & 20 & 0.40\% & 296 \\ 
        8 & 40 & 5.10\% & 3,736 \\ 
        8 & 60 & 6.90\% & 5,024 \\ 
        8 & 80 & 7.20\% & 5,232 \\ 
        12 & 40 & 5.70\% & 4,200 \\ 
        12 & 60 & 6.00\% & 4,440 \\ 
        12 & 80 & 4.10\% & 2,964 \\ 
        16 & 40 & 2.60\% & 1,920 \\ 
        16 & 60 & 5.60\% & 4,128 \\ 
        20 & 60 & 6.50\% & 4,760 \\ 
        \hline
    \end{tabular}
    \label{CG Problem Classes and the Associated  Generated NN Instances}
\end{table}

\section{Hyperparameter Tuning}
\label{Hyperparameter tuning appendix}
The considered hyperparameters are: 
\begin{enumerate}
    \item Model dimension ($d$): Determines the embedding dimension of the model and the size of the subsequent layers after the embedding layer. 
    \item Number of attention heads ($h$): In the multi-head attention sublayers, the input is split into multiple heads, with each attending to different parts of the input. The number of heads determines the level of parallelization and the model's ability to capture different types of interconnections between elements.
    \item Number of encoder and decoder layers: Defines the number of layers in the encoderr of layers allows the model to capture more complex features but increases computational cost.
    \item Dropout rate: A regularization technique that randomly deactivates a certain percentage of neurons during training, reducing the model's reliance on specific connections and preventing overfitting~\citep{srivastava2014dropout}.
    \item  $L_2$ weight decay: Adds a penalty term to the loss function based on the magnitude of the weights to prevent overfitting by encouraging smaller weights and reduce model complexity.
    \item Batch size: The number of training examples processed together in each iteration of the training procedure. Larger batch sizes can provide a smoother gradient estimate but require more memory and computational resources.
    \item Learning rate: The learning rate, denoted by $\alpha$ in the Adam optimizer ~\citep{kingma2014adam}, controls the step size during gradient descent optimization. It determines how quickly the model updates its parameters based on the gradients. Choosing an appropriate learning rate is crucial for achieving good convergence during training.
\end{enumerate}

\begin{landscape}  
\begin{figure}[p] 
        \caption{Validation Loss Vs. Epoch. Each line corresponds to a hyperparameter configuration.}
    \label{fig:LossEpochRayTune}
    \includegraphics[width=\paperwidth]{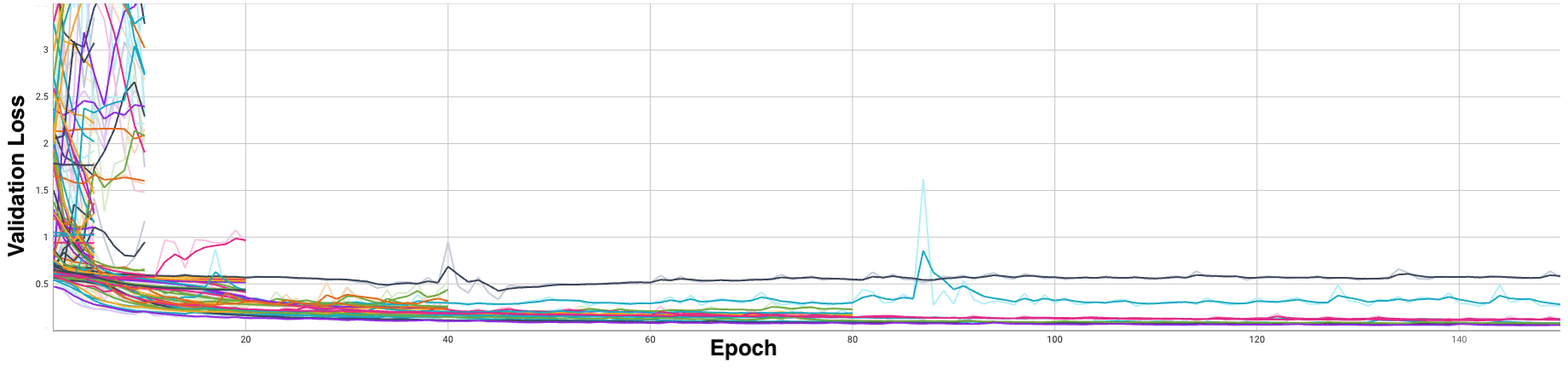}
    \vspace{3cm}
\caption{Validation Accuracy Vs. Epoch. Each line corresponds to a hyperparameter configuration.} \label{fig:AccuracyEpochRayTune}
   {\includegraphics[width=\paperwidth]{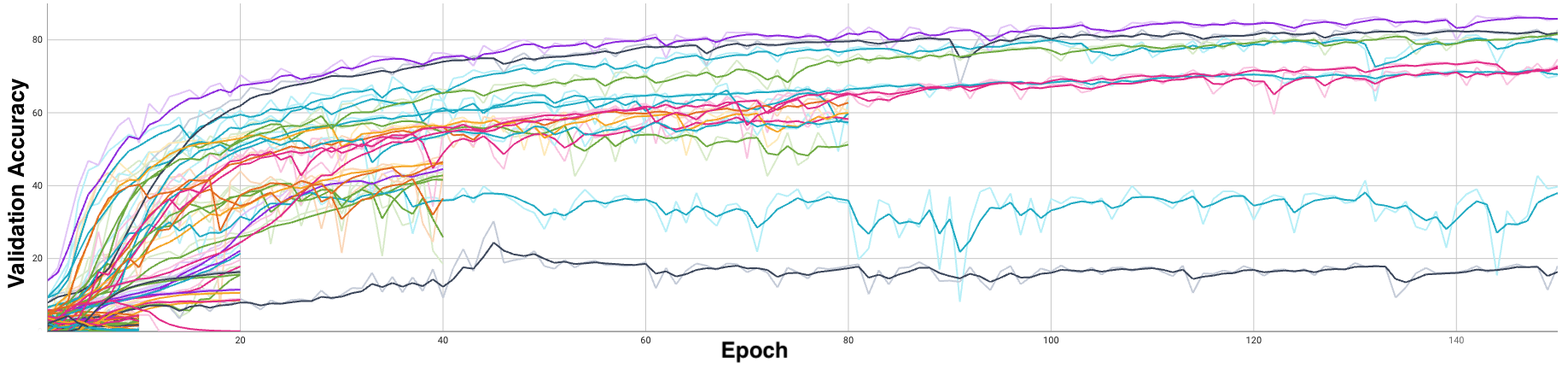}}
\end{figure}
\end{landscape}

We report the results of 100 hyperparameter configurations and their associated validation loss and accuracy in 
Table ~\ref{Evaluated HyperParameter Tuning Configurations and Validation performances}. Figures \ref{fig:AccuracyEpochRayTune} and \ref{fig:LossEpochRayTune} show the results of the hyperparameter tuning trials, including the validation loss and accuracy over 150 training epochs. Early stopped trials are represented by lines that terminate before reaching the maximum number of epochs. These trials exhibit a plateau in the validation loss, suggesting further training is unlikely to improve results.

{\small
\begin{longtable}{|p{1cm}|p{5cm}|p{1cm}|p{1.9cm}|}
    \caption{Evaluated Hyperparameter Tuning Configurations and Validation Accuracy}
    \label{Evaluated HyperParameter Tuning Configurations and Validation performances}\\
    \hline
        \textbf{Trial} & \textbf{Hyperparameter Config} & \textbf{Loss} & \textbf{Accuracy \%} \\ \hline
      \textbf{1} & (32, 4, 6, 6, 0.8, 0.001, 32, 0.0001) & 0.57 & 17.96 \\ \hline
        \textbf{2} & (32, 8, 3, 3, 0.8, 0.001, 16, 0.001) & 0.26 & 39.68 \\ \hline
        \textbf{3} & (64, 4, 3, 6, 0.0, 0.1, 16, 0.001) & 0.93 & 0 \\ \hline
        \textbf{4} & (16, 4, 2, 3, 0.3, 0.001, 16, 1e-05) & 0.66 & 0 \\ \hline
        \textbf{5} & (64, 8, 3, 2, 0.5, 1e-05, 16, 1e-05) & 0.57 & 7.39 \\ \hline
        \textbf{6} & (64, 2, 3, 6, 0.0, 0.1, 32, 0.01) & 3.98 & 0 \\ \hline
        \textbf{7} & (32, 4, 3, 3, 0.5, 0.1, 16, 1e-05) & 0.76 & 0 \\ \hline
        \textbf{8} & (32, 2, 2, 1, 0.3, 0.1, 32, 0.01) & 1.17 & 0.55 \\ \hline
        \textbf{9} & (64, 4, 2, 2, 0.0, 1e-05, 32, 0.0001) & 0.09 & 79.39 \\ \hline
        \textbf{10} & (16, 8, 1, 1, 0.8, 1e-05, 16, 0.0001) & 0.58 & 5.29 \\ \hline
        \textbf{11} & (32, 2, 6, 3, 0.5, 0.1, 16, 0.01) & 3.98 & 0 \\ \hline
        \textbf{12} & (16, 8, 3, 6, 0.3, 0.0, 32, 0.01) & 2.37 & 0 \\ \hline
        \textbf{13} & (32, 4, 3, 3, 0.0, 0.0, 32, 0.0001) & 0.07 & 82.19 \\ \hline
        \textbf{14} & (64, 2, 3, 1, 0.8, 0.0, 32, 0.001) & 1.57 & 5.14 \\ \hline
        \textbf{15} & (32, 8, 3, 3, 0.3, 0.1, 32, 0.0001) & 0.66 & 0 \\ \hline
        \textbf{16} & (16, 8, 3, 2, 0.0, 1e-06, 32, 0.01) & 2.27 & 0 \\ \hline
        \textbf{17} & (32, 2, 2, 1, 0.0, 0.01, 32, 1e-05) & 0.63 & 0 \\ \hline
        \textbf{18} & (16, 2, 3, 1, 0.8, 0.0, 16, 0.01) & 3.98 & 2.87 \\ \hline
        \textbf{19} & (16, 8, 3, 6, 0.0, 0.0, 16, 1e-05) & 0.6 & 0 \\ \hline
        \textbf{20} & (32, 2, 6, 1, 0.8, 0.1, 32, 0.001) & 0.64 & 0 \\ \hline
        \textbf{21} & (64, 4, 6, 1, 0.5, 0.0, 16, 0.01) & 3.98 & 1.83 \\ \hline
        \textbf{22} & (32, 8, 1, 3, 0.3, 1e-06, 32, 0.01) & 1.75 & 0 \\ \hline
        \textbf{23} & (32, 8, 2, 6, 0.5, 1e-06, 16, 0.01) & 3.49 & 0.78 \\ \hline
        \textbf{24} & (64, 4, 1, 6, 0.5, 0.01, 16, 0.001) & 0.94 & 0 \\ \hline
        \textbf{25} & (64, 4, 6, 1, 0.3, 1e-06, 32, 1e-05) & 0.53 & 10.66 \\ \hline
        \textbf{26} & (64, 8, 2, 1, 0.3, 0.0, 16, 1e-05) & 0.51 & 11.56 \\ \hline
        \textbf{27} & (32, 2, 6, 1, 0.5, 1e-06, 32, 0.001) & 0.22 & 52.19 \\ \hline
        \textbf{28} & (16, 2, 2, 1, 0.8, 0.001, 32, 1e-05) & 2.68 & 2.78 \\ \hline
        \textbf{29} & (64, 4, 3, 3, 0.0, 0.0, 32, 1e-05) & 0.43 & 16.63 \\ \hline
        \textbf{30} & (32, 8, 3, 3, 0.5, 0.0, 32, 1e-05) & 0.67 & 0 \\ \hline
        \textbf{31} & (16, 4, 2, 6, 0.0, 1e-06, 32, 0.0001) & 0.11 & 73.06 \\ \hline
        \textbf{32} & (16, 2, 6, 6, 0.3, 0.001, 32, 1e-05) & 1.12 & 0 \\ \hline
        \textbf{33} & (16, 4, 6, 2, 0.8, 0.1, 16, 0.001) & 0.67 & 0 \\ \hline
        \textbf{34} & (16, 8, 3, 3, 0.0, 0.01, 16, 0.001) & 0.91 & 0 \\ \hline
        \textbf{35} & (64, 4, 1, 2, 0.3, 0.1, 16, 0.0001) & 0.63 & 0 \\ \hline
        \textbf{36} & (32, 4, 2, 6, 0.8, 1e-05, 16, 1e-05) & 1.52 & 0 \\ \hline
        \textbf{37} & (16, 4, 1, 6, 0.0, 1e-05, 32, 1e-05) & 0.84 & 0 \\ \hline
        \textbf{38} & (16, 4, 2, 6, 0.5, 0.01, 16, 0.01) & 1.05 & 0 \\ \hline
        \textbf{39} & (64, 8, 6, 1, 0.5, 0.0, 16, 1e-05) & 0.55 & 8.66 \\ \hline
        \textbf{40} & (32, 4, 2, 6, 0.5, 0.001, 32, 0.01) & 4.05 & 1.93 \\ \hline
        \textbf{41} & (64, 4, 1, 3, 0.3, 0.0, 32, 0.0001) & 0.24 & 43.48 \\ \hline
        \textbf{42} & (64, 2, 3, 3, 0.5, 0.001, 16, 0.01) & 3.98 & 1.87 \\ \hline
        \textbf{43} & (16, 4, 3, 1, 0.0, 0.1, 16, 0.0001) & 0.6 & 0 \\ \hline
        \textbf{44} & (32, 4, 1, 6, 0.3, 0.1, 16, 0.0001) & 0.65 & 0 \\ \hline
        \textbf{45} & (16, 2, 2, 3, 0.5, 1e-06, 16, 0.001) & 0.19 & 57.55 \\ \hline
        \textbf{46} & (32, 8, 3, 1, 0.8, 0.0, 16, 1e-05) & 2.11 & 0 \\ \hline
        \textbf{47} & (64, 8, 3, 2, 0.5, 0.001, 32, 0.0001) & 0.33 & 24.74 \\ \hline
        \textbf{48} & (32, 8, 3, 6, 0.3, 0.01, 16, 0.0001) & 0.42 & 17.9 \\ \hline
        \textbf{49} & (32, 2, 2, 3, 0.5, 0.1, 16, 0.01) & 1.7 & 0.01 \\ \hline
        \textbf{50} & (16, 8, 3, 1, 0.0, 1e-06, 16, 1e-05) & 0.58 & 0 \\ \hline
        \textbf{51} & (32, 2, 3, 3, 0.3, 1e-05, 32, 0.0001) & 0.35 & 23.89 \\ \hline
        \textbf{52} & (16, 2, 1, 3, 0.8, 0.0, 16, 1e-05) & 1.48 & 0.31 \\ \hline
        \textbf{53} & (32, 8, 3, 6, 0.3, 1e-05, 32, 0.01) & 3.98 & 0 \\ \hline
        \textbf{54} & (16, 8, 2, 3, 0.0, 0.1, 16, 0.0001) & 0.6 & 0 \\ \hline
        \textbf{55} & (16, 8, 3, 1, 0.0, 0.001, 16, 1e-05) & 0.6 & 0 \\ \hline
        \textbf{56} & (16, 4, 2, 2, 0.8, 0.0, 16, 0.0001) & 0.67 & 3.11 \\ \hline
        \textbf{57} & (64, 8, 6, 1, 0.3, 1e-05, 16, 0.0001) & 0.07 & 82.6 \\ \hline
        \textbf{58} & (64, 8, 3, 3, 0.5, 0.1, 32, 0.01) & 1.02 & 0 \\ \hline
        \textbf{59} & (64, 8, 2, 2, 0.0, 0.01, 16, 1e-05) & 0.54 & 8.9 \\ \hline
        \textbf{60} & (16, 2, 1, 6, 0.3, 1e-05, 16, 0.001) & 0.17 & 60.45 \\ \hline
        \textbf{61} & (64, 4, 2, 2, 0.5, 0.0, 32, 0.0001) & 0.23 & 45.72 \\ \hline
        \textbf{62} & (32, 8, 6, 3, 0.5, 1e-06, 32, 1e-05) & 0.65 & 0 \\ \hline
        \textbf{63} & (32, 8, 1, 2, 0.8, 1e-05, 32, 0.001) & 0.23 & 48.1 \\ \hline
        \textbf{64} & (16, 8, 6, 1, 0.3, 1e-05, 16, 0.01) & 3.98 & 1.88 \\ \hline
        \textbf{65} & (64, 2, 3, 3, 0.0, 1e-05, 32, 0.01) & 2.21 & 0 \\ \hline
        \textbf{66} & (16, 4, 2, 1, 0.0, 0.0, 16, 0.0001) & 0.11 & 74.69 \\ \hline
        \textbf{67} & (32, 8, 2, 6, 0.5, 1e-06, 32, 0.01) & 4.06 & 0.21 \\ \hline
        \textbf{68} & (32, 2, 3, 3, 0.3, 0.1, 16, 1e-05) & 0.72 & 0 \\ \hline
        \textbf{69} & (64, 4, 1, 3, 0.8, 1e-05, 16, 0.001) & 0.52 & 18.46 \\ \hline
        \textbf{70} & (64, 2, 6, 3, 0.5, 0.001, 16, 0.001) & 0.28 & 43.26 \\ \hline
        \textbf{71} & (32, 2, 6, 3, 0.5, 0.1, 16, 0.001) & 1.05 & 0.01 \\ \hline
        \textbf{72} & (64, 2, 6, 3, 0.5, 1e-06, 32, 0.001) & 0.19 & 56.68 \\ \hline
        \textbf{73} & (64, 4, 6, 2, 0.8, 0.001, 16, 0.01) & 3.98 & 1.94 \\ \hline
        \textbf{74} & (64, 8, 2, 3, 0.3, 0.001, 32, 0.0001) & 0.23 & 45.73 \\ \hline
        \textbf{75} & (64, 8, 2, 2, 0.0, 1e-06, 16, 0.0001) & 0.06 & 85.71 \\ \hline
        \textbf{76} & (16, 8, 1, 2, 0.5, 0.001, 32, 0.001) & 0.26 & 41.5 \\ \hline
        \textbf{77} & (16, 4, 2, 3, 0.0, 0.0, 32, 0.01) & 2.16 & 0.21 \\ \hline
        \textbf{78} & (64, 2, 2, 3, 0.3, 0.01, 16, 1e-05) & 0.59 & 0 \\ \hline
        \textbf{79} & (32, 2, 6, 3, 0.3, 0.01, 16, 0.0001) & 0.57 & 5.13 \\ \hline
        \textbf{80} & (16, 2, 3, 3, 0.3, 0.001, 32, 1e-05) & 0.94 & 0 \\ \hline
        \textbf{81} & (64, 2, 2, 6, 0.0, 1e-05, 16, 0.001) & 0.91 & 0 \\ \hline
        \textbf{82} & (16, 4, 1, 6, 0.5, 0.1, 16, 0.01) & 1.14 & 0.28 \\ \hline
        \textbf{83} & (32, 2, 3, 2, 0.0, 1e-05, 16, 1e-05) & 0.44 & 15.98 \\ \hline
        \textbf{84} & (16, 8, 2, 3, 0.5, 0.01, 32, 0.001) & 0.6 & 0 \\ \hline
        \textbf{85} & (16, 8, 1, 3, 0.8, 1e-05, 16, 0.01) & 1.78 & 0 \\ \hline
        \textbf{86} & (32, 8, 3, 1, 0.3, 1e-06, 16, 0.0001) & 0.13 & 70.14 \\ \hline
        \textbf{87} & (64, 8, 1, 3, 0.5, 0.01, 32, 0.0001) & 0.59 & 0 \\ \hline
        \textbf{88} & (64, 4, 1, 2, 0.8, 0.0, 32, 0.01) & 2.65 & 0.01 \\ \hline
        \textbf{89} & (16, 2, 1, 1, 0.3, 1e-05, 16, 0.0001) & 0.57 & 0 \\ \hline
        \textbf{90} & (32, 2, 3, 1, 0.5, 1e-05, 32, 0.01) & 3.98 & 5.22 \\ \hline
        \textbf{91} & (32, 4, 1, 1, 0.8, 0.01, 16, 0.0001) & 0.61 & 0 \\ \hline
        \textbf{92} & (64, 4, 3, 6, 0.5, 0.001, 16, 0.01) & 2.33 & 0.72 \\ \hline
        \textbf{93} & (32, 8, 2, 2, 0.3, 1e-05, 16, 0.001) & 0.15 & 63.35 \\ \hline
        \textbf{94} & (64, 2, 2, 3, 0.5, 0.001, 32, 0.0001) & 0.37 & 19.63 \\ \hline
        \textbf{95} & (64, 4, 1, 3, 0.5, 0.01, 32, 0.01) & 1.38 & 0.73 \\ \hline
        \textbf{96} & (32, 2, 6, 6, 0.5, 0.01, 16, 1e-05) & 0.63 & 0 \\ \hline
        \textbf{97} & (16, 2, 3, 2, 0.8, 0.001, 16, 0.01) & 1.99 & 3.47 \\ \hline
        \textbf{98} & (32, 2, 1, 6, 0.0, 1e-06, 32, 0.0001) & 0.15 & 64.16 \\ \hline
        \textbf{99} & (16, 8, 1, 6, 0.8, 1e-05, 16, 0.01) & 3.36 & 0 \\ \hline
        \textbf{100} & (32, 4, 1, 3, 0.8, 0.01, 32, 1e-05) & 1.92 & 0 \\ \hline
\end{longtable}
}
    \centering

\end{document}